\documentclass[10pt,twocolumn,letterpaper]{article}

\usepackage{cvpr}              

\usepackage{booktabs}    
\usepackage{multirow}    
\usepackage{colortbl}    

\usepackage{subcaption}    

\usepackage[ruled,vlined]{algorithm2e}   
\usepackage{xcolor} 

\usepackage{float}         

\definecolor{cvprblue}{rgb}{0.21,0.49,0.74}
\definecolor{myorange}{rgb}{0.98,0.49,0.169}
\definecolor{mycyan}{rgb}{0.0078,0.9451,0.9529}
\usepackage[pagebackref,breaklinks,colorlinks,allcolors=cvprblue]{hyperref}

\def\paperID{} 
\def\confName{CVPR}
\def\confYear{2026}

\newcommand{\name}[0]{KVSmooth\xspace}
\DeclareMathOperator*{\argmax}{argmax}

\DeclareMathOperator*{\chairs}{CHAIR_S}
\DeclareMathOperator*{\chairsr}{CHAIR_{SR}}
\DeclareMathOperator*{\chairi}{CHAIR_I}
\DeclareMathOperator*{\chair}{CHAIR}

\title{\name: Mitigating Hallucination in Multi-modal Large Language Models through Key-Value Smoothing}

\author{
Siyu Jiang$^{1,*}$ \quad
Feiyang Chen$^{1,*}$ \quad
Xiaojin Zhang$^{1}$ \quad
Kun He$^{1,\dagger}$\\
$^{1}$
Huazhong University of Science and Technology\\
{\tt\small
\{jiangsiyu, chenfeiyang, xiaojinzhang, brooklet60\}@hust.edu.cn}
}

\begin{document}
\maketitle
\def\thefootnote{*}\footnotetext{Equal contribution.}
\def\thefootnote{$\dagger$}\footnotetext{Corresponding author.}
\setcounter{footnote}{0}

\begin{abstract}
Despite the significant progress of Multi-modal Large Language Models (MLLMs) across diverse tasks, hallucination, which corresponds to the generation of visually inconsistent objects, attributes, or relations, remains a major obstacle to their reliable deployment. Unlike pure language models, MLLMs must ground their generation process in visual inputs; 
However, existing models often suffer from semantic drift during decoding, causing outputs to diverge from visual facts as the sequence length increases.
To address this, we propose \name, a training-free,  plug-and-play method that mitigates hallucination by performing attention–entropy–guided adaptive smoothing on hidden states. Specifically, \name applies an exponential moving average (EMA) to both keys and values in the KV-Cache while dynamically quantifying the sink degree of each token through its attention distribution entropy to adaptively adjust the smoothing strength. 
Unlike computationally expensive retraining or contrastive decoding methods, 
\name operates efficiently during inference without additional training or model modification. 
Extensive experiments demonstrate that \name significantly reduces hallucination ($\mathit{CHAIR}_{S}$ from $41.8 \rightarrow 18.2$) while improving overall performance ($F_1$ score from $77.5 \rightarrow 79.2$), achieving higher precision and recall simultaneously, whereas prior methods often sacrifice one for the other, thereby validating the effectiveness and generality of our method.
\end{abstract}
\section{Introduction}
\label{sec:intro}
Multi-modal Large Language Models (MLLMs)~\cite{llava, minigpt4, instructblip2, bai2025qwen2} have achieved remarkable success across various vision–language tasks, including image captioning~\cite{chair}, visual question answering \cite{guo2023images}, and multi-modal dialogue \cite{DBLP:conf/acl/SunWXZ0HXZGJ22}. Typically, these models adopt an \emph{encoder–alignment–decoder} framework: a vision encoder extracts visual representations, a lightweight alignment module (e.g., Q-Former \cite{li2023blip} or an MLP) maps them into the linguistic space, and a pretrained LLM 
— further trained with the alignment module — performs reasoning and generation. 
This architecture effectively integrates 
visual and textual representations, 
enabling coherent and contextually grounded text generation while preserving the LLM's inherent 
reasoning ability and 
maintaining computational efficiency. 
Despite these advances, MLLMs often over-rely on linguistic priors, resulting in \textbf{multi-modal hallucination}, the generation of content misaligned with the visual input~\cite{huang2025survey, liu2024survey}. Although numerous strategies have been explored to mitigate this issue~\cite{jiang2024hallucination, chen2025ict, chen2024multi, cho2025you, DBLP:conf/cvpr/FaveroZTCPASS24},  hallucination remains a 
major challenge to achieving trustworthy multi-modal reasoning.

To address this issue, we examine its origin from a generation dynamics perspective. Unlike LLMs that rely solely on linguistic context, MLLMs must tightly align each generated token with visual evidence. This creates a fundamental tension: as decoding progresses, the influence of early visual tokens often decays within hidden representations, weakening visual grounding. The resulting  \textbf{semantic drift} leads the model to produce text that gradually diverges from the image content. We identify two primary challenges:

\begin{enumerate}
    \item \textbf{Long-term visual dependency decay.} Visual cues gradually fade over long decoding horizons, leading the generated text to drift away from the image context.
    \item \textbf{Cumulative semantic drift.} Small early-generation 
    inaccuracies accumulate over time, progressively amplifying the gap between 
    the generated description and the visual content.
\end{enumerate}

Correspondingly, we have the following three findings:
First, through systematic analysis, we observe a clear distinction in logit dynamics: the mean and variance of \textit{true-object} logits exhibit a \textbf{monotonic decrease} over the decoding steps, whereas those of \textit{hallucinated-object} logits display a steady increase.
We then introduce attention row-entropy to quantify a token’s sink strength, offering a real-time alternative to attention column-sum. Our analysis reveals that tokens attracting disproportionate attention are characterized by high row-entropy, indicating an over-smoothed aggregation that dilutes critical visual information. 
Consequently, \textbf{row-entropy is positively correlated with hallucination probability}, explaining why sink tokens often act as precursors to hallucination. Macroscopically, sink tokens emerge at major semantic transitions, where the model averages across previous states to leap into a new semantic space. In multi-modal reasoning, such aggressive transitions can easily deviate from the visual context, leading to hallucinations.

Inspired by these findings, we propose \textbf{\name}, a lightweight, training-free, and plug-and-play solution. Our method introduces two key components:
(1) \textbf{Exponential Moving Average (EMA) Smoothing on KV-Cache}, which applies token-wise EMA smoothing to Key and Value caches to suppress abrupt state changes and curb the variance explosion of hallucinatory logits;
(2) \textbf{Entropy-guided coefficient adaptation}, which dynamically increases the smoothing strength for high-entropy sink tokens, further mitigates their impact on generation and reduces hallucination.

\name offers a novel and effective approach to mitigate hallucinations in MLLMs by stabilizing hidden-state dynamics and provides a reliable and practical foundation for safer multi-modal generation. The main contributions are as follows:
\begin{itemize}
    \item We introduce the concept of \emph{sink degree}, a novel metric defined via attention row-entropy that enables continuous, real-time identification of tokens prone to hallucination.
    \item We propose \name, a lightweight and training-free method that employs adaptive EMA smoothing on the KV-Cache. By dynamically adjusting the smoothing strength based on the sink degree, \name effectively suppresses hidden-state perturbations that lead to hallucinations without compromising generation efficiency.
    \item Extensive experiments across multiple benchmarks and MLLMs validate the effectiveness and generality of \name, demonstrating significant gains in reducing hallucinations while maintaining comprehensiveness. 
\end{itemize}


\section{Related Work}
\label{sec:related work}

\subsection{Hallucinations in MLLMs}

Existing efforts to mitigate hallucination in MLLMs fall into two main categories. The first fine-tunes models to align with human preferences or ground-truth annotations~\cite{xiao2025detecting, yang2025mitigating, wu2025antidote, jiang2024hallucination, lyu2024alleviating, liang2025mole}. For example, POVID~\cite{povid} uses reinforcement learning with AI-generated feedback and annotated instructions as preferred responses, effectively reducing hallucination through direct preference optimization. However, such methods demand substantial training data and computational resources.


The second stream avoids parameter updates and instead addresses hallucination through decoding strategies~\cite{vcd, park2025convis, DBLP:conf/iclr/GhoshEKTNJM25, li2025hidden, DBLP:conf/iclr/Huo0ZWCZ25,itad} or attention reweighting mechanisms~\cite{visualsink, pai, Middlelayer, DBLP:conf/iclr/TangHLSYL25, sparc, DBLP:conf/iclr/YangL0X25,yang2025mitigating}. Visual contrastive decoding~\cite{vcd} amplifies and removes hallucinated distributions using noise-augmented views. Follow-up studies further extend contrastive decoding by expanding contrastive sets with diffusion-generated images~\cite{park2025convis}, introducing description-conditioned prefixes~\cite{DBLP:conf/iclr/GhoshEKTNJM25}, or adopting layer-aware contrastive strategies~\cite{itad}, all achieving consistent hallucination reduction without retraining.

Attention reweighting methods aim to rebalance cross-modal attention to mitigate modality imbalance. For instance, PAI~\cite{pai} enhances attention to visual regions, while MiddleLayer~\cite{Middlelayer} adjusts visual attention by integrating information across multiple heads. However, both approaches tend to suppress correctly described objects along with hallucination. To address this, SPARC~\cite{sparc} tracks cross-step attention gaps to detect critical visual tokens and selectively reinforces those whose influence increases over time, thereby slightly reducing hallucinations while maintaining high recall. Meanwhile, TAME~\cite{DBLP:conf/iclr/TangHLSYL25} identifies hallucination as stemming from polarized variance in attention spectra, where extreme eigenvalues cause over-anchoring on a few tokens. They rescale Query–Key variances to restore attention balance. PruneHal~\cite{sun2025prunehal} reduces hallucinations by pruning redundant visual tokens on KV-Cache, which reallocates the model's attention to important visual tokens. Overall, attention misalignment reveals MLLMs' chronic under-utilization of visual cues during deep generation.

\subsection{Attention Sink in MLLMs}
The attention sink phenomenon was first identified in StreamingLLM~\cite{DBLP:conf/iclr/XiaoTCHL24}, where LLMs consistently assigned disproportionately high attention to the beginning-of-sequence (BOS) token.
A recent study~\cite{DBLP:conf/iclr/GuPDLZD0L25} provides a geometric explanation: key of the first token lies on a distinct manifold, resulting in small angles with most queries and causing an attention sink, where attention disproportionately focuses.

OPERA~\cite{opera} generalized this concept beyond BOS, identifying semantically redundant “aggregation tokens” that persistently receive maximal attention across decoding steps. These textual sink tokens absorb substantial attention resources while contributing little factual grounding, and their over-activation correlates strongly with hallucinations—extending the sink notion from a positional artifact to a broader linguistic behavior in auto-regressive decoding.

To mitigate sink-induced hallucinations, methods such as OPERA~\cite{opera} employ an over-trust logit penalty and retrospection-allocation to revisit and re-evaluate sink tokens.  AttnReal~\cite{attnreal} redistributes attention from textual sinks toward visual evidence, while FarSight~\cite{tang2025seeing} reserves attention for future context to reduce sink dominance. Though differing in strategy, these approaches share the goal of suppressing or reallocating attention away from sink tokens.

However, existing work primarily focuses on reducing the occurrence of sink tokens or diminishing the attention allocated to the most dominant sink tokens, without explaining why sink tokens trigger hallucinations. In this study, we reveal that aggregation tokens, while integrating context, can distort internal representations during fusion and directly induce hallucinations. Correspondingly, we propose to regularize and correct the hidden states of such tokens, aiming to suppress hallucinations by stabilizing their semantic contributions and enhancing multi-modal factual grounding.




\section{Observation}
In this section, we begin by formulating MLLM generation, and then describe our three key observations in detail.

\label{observations}
\begin{figure}[t]
    \centering
    \includegraphics[width=0.4\textwidth]{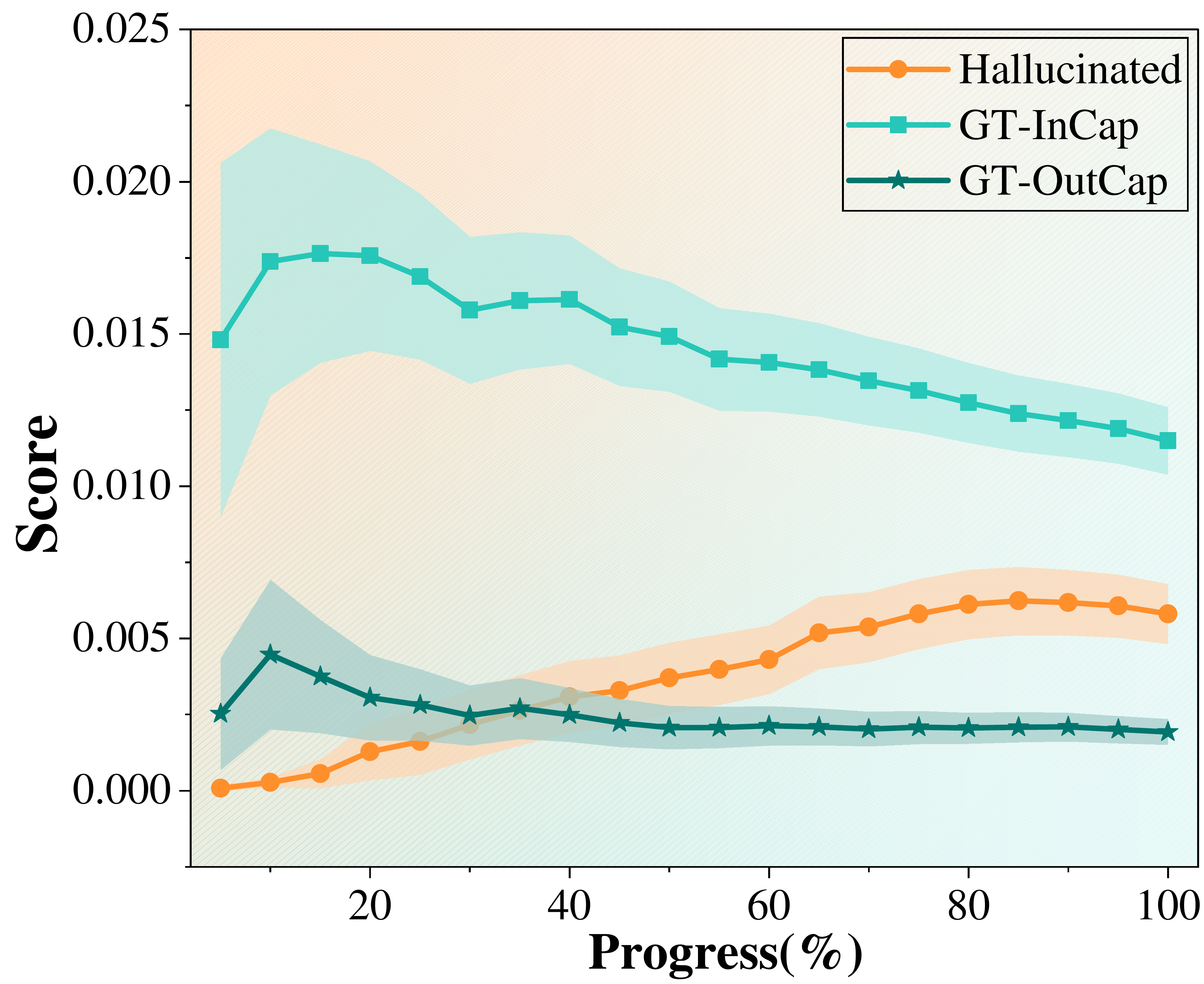}
    \caption{
    Variation of object logit scores during caption generation.
    We analyze 200 images and compute the average score of each object category across different generation stages.
    Objects are categorized into three groups:
    (1) \textbf{GT-InCap} - objects appearing in both the image and the generated caption,
    (2) \textbf{GT-OutCap} - objects present in the image but missing from the caption, and
    (3) \textbf{Hallucinated} - objects mentioned in the caption but absent from the image.
    The y-axis denotes the average logit score of each object group, while the x-axis represents the generation progress.
    Each caption is divided into twenty stages by token count, where each stage includes all tokens 
    up to that point.
    The mean and variance are computed for each stage and 95\% confidence intervals are reported.
    }
    \label{fig1}
\end{figure}

\begin{figure}[h!]
    \centering
    \includegraphics[width=0.45\textwidth]{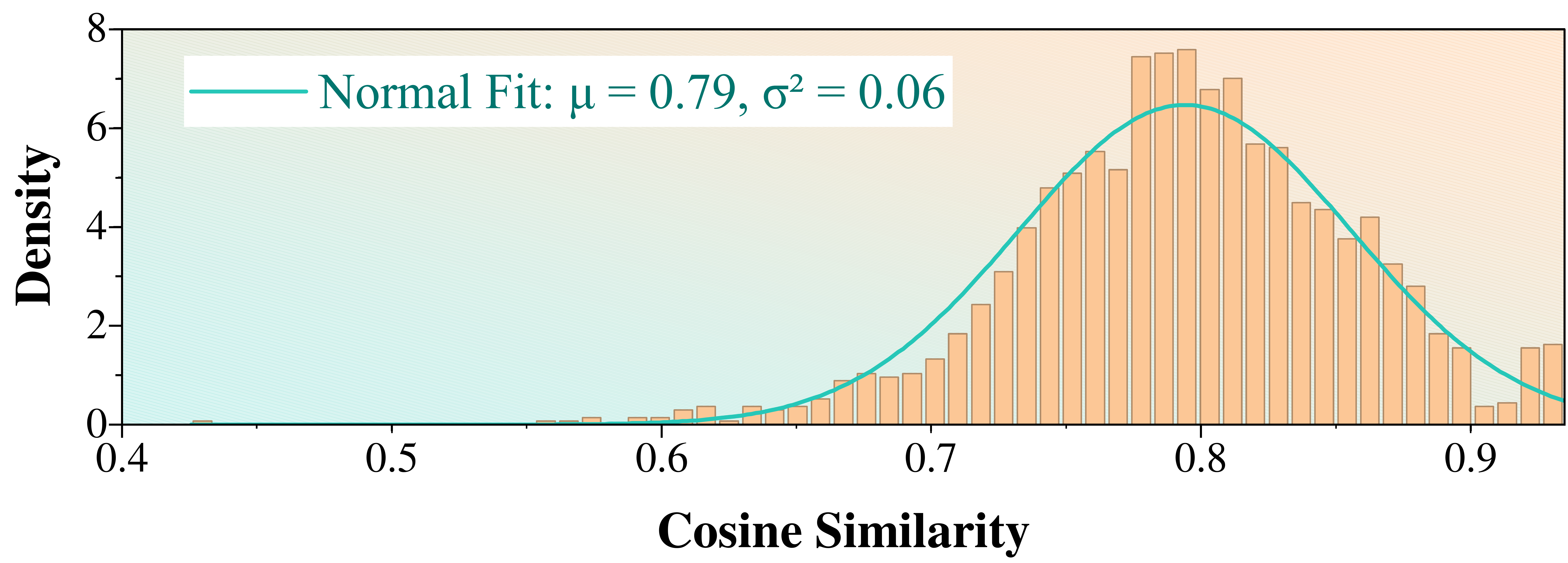}
    \caption{
    Distribution of cosine similarity between attention row-entropy and column-sum across all layers during the generation process.
    The similarity values exhibit a precise unimodal distribution centered around 0.79 with low variance, indicating a stable and strong positive correlation between attention row-entropy and column-sum.
    }
    \label{fig2}
\end{figure}
\begin{figure}[h!]
    \centering
    \includegraphics[width=0.45\textwidth]{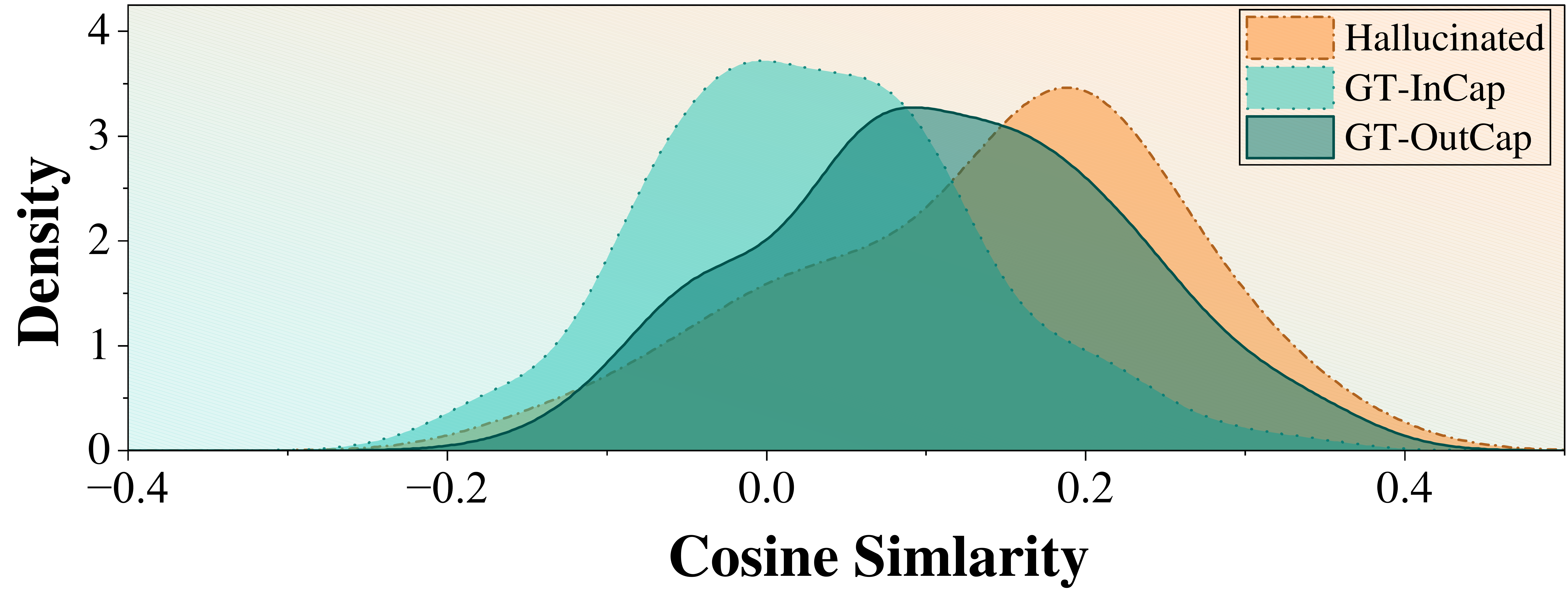}
    \caption{
    Distribution of cosine similarity between logit ranking and attention row-entropy across object types in 200 images.
    We compute the cosine similarity between row-entropy and ranking scores for three object categories. Hallucinated objects exhibit the highest similarity, indicating that greater row-entropy correlates with stronger hallucination tendencies, whereas genuine objects (GT-InCap and GT-OutCap) show lower or slightly negative correlations.
    }
    \label{fig3}
\end{figure}

\subsection{Formulation of MLLMs Generation}
\label{sec:fomulation of mmlms generation}
MLLMs process both images and text. The image is transformed into a sequence of visual tokens $x^v = \{x_0, x_1, ..., x_{N-1}\}$, where $N$ is typically fixed, while the input text is tokenized into $x^p = \{x_N, x_{N+1}, ..., x_{N+M-1}\}$. These visual and text tokens are concatenated to form the initial input sequence for the decoder.

Generation proceeds auto-regressively. After generating $k$ tokens, the decoder input is of length $L = N + M + k$, denoted as $\{x_0, x_1, ..., x_{L-1}\}$. The model then computes the probability distribution for the next token at position $t = L$.

Let $Q^{(l,h)}_t$, $K^{(l,h)}_j$, $V^{(l,h)}_j \in \mathbb{R}^d$ represent the query, key, and value vectors of head $h$ in layer $l$, where $t = L$ is the current token position, and $j = 0, 1, ..., L-1$ indexes the context. The attention score is:
\begin{equation}
A^{(l,h)}_{t,j} = \frac{\bigl(Q^{(l,h)}_t\bigr)^\top K^{(l,h)}_j}{\sqrt{d}},
\end{equation}
and is normalized via softmax over $j$:
\begin{equation}
\alpha^{(l,h)}_{t,j} = \frac{\exp\bigl(A^{(l,h)}_{t,j}\bigr)}{\sum_{n=0}^{L-1} \exp\bigl(A^{(l,h)}_{t,n}\bigr)}.
\end{equation}
The output of head $h$ is:
\begin{equation}
o^{(l,h)}_t = \sum_{j=0}^{L-1} \alpha^{(l,h)}_{t,j} V^{(l,h)}_j.
\end{equation}
Outputs from all $H$ heads are concatenated and projected to form layer-$l$'s hidden state $h^l_t$, which is passed to the next block.

Finally, the hidden state of the last layer is projected to vocabulary logits. The next-token probability is:
\begin{equation}
p_\theta(y_t \mid v, x, y_{<t}) \propto \exp(\text{logit}_\theta(y_t \mid v, x, y_{<t})),
\end{equation}
where $\text{logit}_\theta(y_t \mid v, x, y_{<t})$ is the unnormalized logit for $y_t$. The next token is sampled from this distribution, and this process repeats.

\subsection{Divergent Logit Dynamics of Hallucinated vs. Non-Hallucinated Objects}\label{ob1}
We first examine how the logit distributions of different object categories evolve during decoding. 
As shown in Figure~\ref{fig1}, Ground Truth (GT) objects - comprising both GT-InCap and GT-OutCap - exhibit a monotonic decline in mean logit while their variance stabilizes, indicating that the model's focus on visually grounded cues gradually weakens, 
causing their scoring advantage to fade over time.
In contrast, hallucinated objects display rising mean logits accompanied by slightly increasing variance. 
This divergence suggests that spurious candidates accumulate instability in hidden representations, making them more likely to be erroneously sampled in later steps.
Thus, preserving a stable and dominant signal for GT objects while suppressing the anomalous amplification of hallucinated ones is essential for effective hallucination mitigation.

\subsection{
Row-Entropy Affinity and Attention Sinks
} \label{ob2}

Previous work~\cite{opera} introduced a column-wise attention score to identify aggregation tokens. However, computing tokens with large attention column-sums requires multiple decoding steps and backtracking, which is computationally inefficient. 
To address this, we propose attention row-entropy as a real-time metric for quantifying a token's sink strength. Inspired by \cite{DBLP:conf/iclr/GuPDLZD0L25}, the BOS token occupies an average angular position in the hidden-state space, naturally exhibiting high cosine similarity with other tokens and, consequently, high attention scores. We hypothesize that aggregation tokens follow a similar pattern.

Formally, we define the attention row-entropy of token $x_t$ as 
\begin{equation}
    \label{eq:row-entropy}
    z^l_{t}= -\dfrac{1}{H}{\displaystyle\sum_{j=0}^{L-1}\sum_{h=0}^{H-1}}     \alpha^{(l,h)}_{t,j}\,\log\!\bigl(\alpha^{(l,h)}_{t,j}+\varepsilon\bigr),
\end{equation} 
which serves as a proxy for its aggregation strength.

As shown in Figure~\ref{fig2}, tokens with high row-entropy - \ie, diffuse attention distributions - tend to produce hidden states that approximate the historical average, lying near the central manifold of past representations. These tokens thus exhibit small angular distances to most hidden states and attract disproportionately high attention in subsequent steps, forming attention sinks.
Compared to aggregation centered on a few key tokens, 
global averaging introduces 
stronger hidden-state distortions that accumulate over time, amplifying hallucinations as subsequent tokens repeatedly attend to these sinks.


\subsection{Entropy-Ranking Coupling in Hallucinated Objects} \label{ob3}
While prior studies have linked hallucinations to sink tokens, the underlying mechanism remains unclear. To elucidate this, we analyze the joint evolution of logit rankings and attention row-entropy for hallucinated versus faithful object tokens during generation.

We observe a weak positive correlation between ranking and row-entropy for hallucinated objects: the more uniformly a token attends (\ie, higher entropy), the higher the logits of hallucinated objects tend to rise. GT objects, however, exhibit the opposite trend. This finding reveals a direct link between attention uniformity and hallucination risk: sink tokens, which average across the entire context, systematically inflate hallucinated object scores.

At a broader level, this behavior originates from pre-trained LLMs, which summarize previously observed entities and continuously introduce new entities during generation.
When fine-tuned into MLLMs, this tendency persists: repeated high-entropy aggregations continuously amplify spurious object scores, allowing them to dominate the output.

\subsection{Summary of Insights} 
In summary, 
our analysis identifies variance propagation via sink tokens as the root cause of object hallucination.  
\textbf{Obs1} 
    reveals that hallucination scores increase while GT scores decay over time.
 \textbf{Obs2} 
    demonstrates that attention row-entropy strongly correlates with aggregation (sink) behavior.
 \textbf{Obs3} 
    establishes the causal coupling between attention row-entropy and hallucination amplification.
Together, 
these findings motivate a variance-regularized state-evolution framework aimed at (1) maintaining GT object dominance and (2) suppressing hallucination-inducing stochastic fluctuations from sink tokens. 

\section{Methodology}
\label{method}

In this section, we begin by 
deriving an Exponential Moving Average (EMA)–based smoothing strategy following \textbf{Obs1}. 
Both theoretical analysis and empirical results show that applying EMA to the key and value is most effective, forming our EMA Smoothing on KV-Cache method (Section~\ref{sec:ema}).
Then, inspired by \textbf{Obs2} and \textbf{Obs3}, we introduce 
an adaptive coefficient selection mechanism guided by attention entropy, which applies stronger smoothing to tokens with higher sink tendencies, termed Entropy-Guided Coefficient Adaptation (Section~\ref{sec:adaptive_coefficient}).




\subsection{EMA Smoothing on KV-Cache}
\label{sec:ema}


As encouraged by \textbf{Obs1}, the mean and variance of the hallucination object logit anomaly increase, indicating that the standard decoding process lacks the $p(h_t | h_{<t})$ constraint.
To keep the decoding trajectory ideally smooth, we formalize the following hypothesis:  
\begin{equation}
    \label{eq:gauss}
    h_t=h_{t-1} + \epsilon_t, ~\epsilon_t\sim N(0, \sigma^2_p).
\end{equation}
Consequently, the transition prior is  
\begin{equation}
\label{eq:priori}
    \begin{split}
    P(h_t | h_{t-1})=N(h_t;h_{t-1},\sigma^2_p)\propto \\
    \exp\left(-\frac{1}{2\sigma^2_p}||h_t - h_{t-1}||^2\right). 
    \end{split}
\end{equation}
Under this prior, the maximum-a-posteriori (MAP) estimate of $h_t$ is given by  
\begin{equation}
\label{eq:map}
\hat{h_t}=\argmax_{h_t}\left[ \underbrace{\log P(o_t|h_t)}_{\text{likelihood}} + \underbrace{\log P(h_t|h_{t-1})}_{\text{smoothed prior}} \right],
\end{equation}
where $o_t$ denotes the raw hidden state observed at step $t$.  
Assuming both terms follow Gaussian distributions,  
\begin{equation}
    \label{eq:insert map}
    \hat{h_t} = \argmax_{h_t}\left[ 
    -\frac{1}{2\sigma^2_o}||o_t - h_{t}||^2 -\frac{1}{2\sigma^2_p}||h_t - h_{t-1}||^2
    \right].
\end{equation}
Taking the derivative of \eqref{eq:insert map} with respect to $h_t$, setting it to zero, and solving for $\hat{h_t}$ yields 
\begin{equation}
    \label{eq:preema}
    \hat{h_t} = \frac{\sigma^2_p}{\sigma^2_p + \sigma^2_o}o_t + \frac{\sigma^2_o}{\sigma^2_p + \sigma^2_o}h_{t-1}.
\end{equation}
Defining  
\begin{equation}
    \lambda_t = \frac{\sigma^2_o}{\sigma^2_p + \sigma^2_o},  
\end{equation}
the estimate simplifies to an exponential moving average (EMA) form:   
\begin{equation}
    \label{eq:ema}
    \hat{h_t} = (1 - \lambda_t)o_t + \lambda_t h_{t - 1}.
\end{equation}
Thus, when the likelihood follows Gaussian in $h_t$, the MAP estimator coincides exactly with an exponential moving average. The detailed derivation can be found in Appendix~\ref{appendix:detail derivation of ema smoothing}.

Motivated by this equivalence, we apply an EMA update directly to the current hidden state $h_t$. This operation constitutes an efficient, closed-form instantiation of the Bayesian-optimal smoother, optimally balancing fidelity to the current observation with temporal consistency.

Another question arises: how exactly should we smooth 
$h_t$, and at which stage should this be applied?
Empirically, we find that simultaneously updating the key $K_t$ and value $V_t$ stored in the preceding KV-Cache, rather than the raw hidden variable itself, could maximally suppress both mean and variance of the logits. Section~\ref{sec:fomulation of mmlms generation} has shown how $K_t$ and $V_t$ yield the hidden state $h_t$.  
By simultaneously smoothing $K_t$ and $V_t$ so that each satisfies the distribution in Eq. \eqref{eq:gauss}, and exploiting the fact that the product of two independent Gaussian distributions remains Gaussian, we conclude that the resulting product $h_t$ also obeys the same distribution.

By operating on the KV-Cache, a larger smoothing coefficient could be applied, achieving the most substantial regularization effect on the logits and consequently yielding the most effective hallucination suppression. The experiments in Section~\ref{subsec: effect of smoothing method} further validate the effectiveness of smoothing $K_t$ and $V_t$, while Appendix~\ref{appendix:layer_sensitivity} investigates the optimal layers for applying such smoothing.

\subsection{
Entropy-Guided Coefficient Adaptation}
\label{sec:adaptive_coefficient}
Prior work has shown that sink tokens are more likely to trigger hallucinations.
As revealed in \textbf{Obs2}, the traditional column-sum metric used to identify sink tokens is positively correlated with our proposed row-entropy: higher row-entropy indicates a greater likelihood that a token acts as a sink token and induces hallucination.
Therefore, it is crucial to attenuate the influence of high-entropy tokens during subsequent generation steps. 

In EMA smoothing, the coefficient directly controls the smoothing strength: a larger coefficient yields stronger smoothing, 
thereby reducing the token’s impact on future updates. 
By adaptively modulating this coefficient, we can suppress high row-entropy tokens.
However, determining the optimal coefficient is non-trivial. 
Since $\sigma^2_o$ is unavailable, the overall smoothing coefficient $\lambda_t$ remains intractable. Using a uniform smoothing coefficient for all tokens may also over-suppress the semantic flow of normal tokens, ultimately degrading recall. 
%

Inspired by \textbf{Obs3}, we believe that different tokens contribute unequally to hallucination based on their row-entropy-ranking. We aim to introduce a scalar that can dynamically provide feedback on each token’s contribution to hallucination during the generation process. 

Specifically, at time step $t$, we first compute the row-entropy $z_t^l$ of token $x_t$ for layer $l$ following Eq.~\eqref{eq:row-entropy}.
We maintain a First-In-First-Out queue $S^l = { z_{t-M+1}^l, \dots, z_t^l }$, where $M$ denotes the maximum length of the queue. The queue preserves the temporal order of tokens. We assume that the growth of sequence length $L$ during generation has a negligible effect on the computation of row-entropy within the queue.
After inserting $z_t^l$ into the queue, we compute its percentile rank $k$, \ie, the count of elements in the queue with row-entropy smaller than $z_t^l$, regardless of their positions in the queue. We then calculate the proportion of these elements within the queue and use this value as the smoothing coefficient $\hat{\lambda}_t^l$ for token $x_t$ at layer $l$.
Formally, $\hat{\lambda}_t^l$ is defined as:
\begin{equation}
    \label{eq:alpha}
    \hat{\lambda}^l_t = \frac{k}{M }\in \left[0,\frac{M-1}{M}\right].
\end{equation}

Furthermore, to preserve information from earlier hidden states while retaining the intrinsic property of each current state, 
we clip extremely large or small values of $\hat{\lambda}^l_t$. 
Concretely, we define a hyperparameter $\lambda_{\text{ref}}$ to guide the truncation and perform the clipping as:
\begin{equation}
\tilde{\lambda}^l_t = \max\left(\lambda_{\text{ref}} - 0.2,\; \min\left(\lambda_{\text{ref}} + 0.2,\; \hat{\lambda}^l_t\right)\right),
\end{equation}
which constrains $\hat{\lambda}^l_t$ within a narrow window around itself, stabilizing generation while preserving representational diversity.

\subsection{The Final \name Method}
\label{sec:final}

We designate a subset of decoder layers in the MLLM as smoothing targets. 
At generation step $t$, for each target layer $l$, we compute the token-specific smoothing strength $\hat{\lambda}_t^{l}$ using attention-row-entropy-based adaptive coefficient selection.
The KV-Cache of token $x_t$ on layer $l$ is then updated via EMA smoothing:
\begin{equation}
    \begin{split}
        \hat{K_t^l} = (1 - \tilde{\lambda}_t^{l})K_t^l + \tilde{\lambda}_t^{l}\space \  K_{t - 1}^l,
        \\
        \hat{V_t^l} = (1 - \tilde{\lambda}_t^{l})V_t^l + \tilde{\lambda}_t^{l}\space \  V_{t - 1}^l.  
        \label{kvsmooth}
    \end{split}
\end{equation}

Applying \name\ effectively suppresses the mean and variance of logit probabilities associated with hallucinatory objects by reducing the emergence of sink tokens. 
By keeping the decoder anchored to visual priors, \name\  prevents abrupt state deviations from image-grounded evidence, thereby mitigating hallucination. 
Further algorithmic details are provided in  Appendix~\ref{appendix:fake}. 
Notably, \name\ is lightweight, training-free, and can be seamlessly integrated in a plug-and-play manner.

    
\section{Experiments}
\label{sec:experiment}

\subsection{Experimental Setup}
\textbf{Baselines.}
We compare our method with five representative training-free approaches.
VCD~\cite{vcd} is a typical contrastive decoding method that enhances the influence of visual information.
OPERA~\cite{opera} mitigates hallucination by reducing sink-token effects through over-trust penalty and retrospection-based decoding.
We also include three attention-redistribution methods: PAI~\cite{pai}, SPARC~\cite{sparc}, and MiddleLayer~\cite{Middlelayer}, which enhance visual attention.

\textbf{Implementation Details.}
We evaluate our method on three representative MLLMs: LLaVA-1.5~\cite{llava}, MiniGPT-4~\cite{minigpt4}, and InstructBLIP~\cite{instructblip2} to assess both effectiveness and generalization. All models adopt their 7B versions. Experiments are conducted on four standard hallucination benchmarks: CHAIR~\cite{chair} for caption hallucination; OPOPE~\cite{opope} for object presence verification; AMBER~\cite{amber} for comprehensive multi-scenario evaluation; and Object HalBench~\cite{chair} evaluates object hallucination by leveraging GPT~\cite{gpt4} to assist in object extraction. All models perform greedy decoding with a maximum of 512 generated tokens. Our method applies EMA smoothing to layers 3–31 across all models, using a fixed FIFO queue of length 15. The reference hyperparameter $\lambda_{\text{ref}}$ is set to 0.9, 0.5, and 0.7 for LLaVA-1.5, MiniGPT-4, and InstructBLIP, respectively, and this configuration is applied consistently across all benchmark evaluations.

\begin{table*}[t]
\centering
\caption{CHAIR performance of different models. Lower $\chairs$ and higher $\text{F}_{1}$ indicate better performance. The best results in each column are highlighted in \textbf{bold}, and the second-best are  \underline{underlined}.}
\renewcommand\arraystretch{1.15}
\setlength{\tabcolsep}{8pt}
\scalebox{0.8}
{
\begin{tabular}{lcccccc}
\toprule
\multirow{2}{*}{Method} & 
\multicolumn{2}{c}{LLaVA-1.5} & 
\multicolumn{2}{c}{MiniGPT-4} & 
\multicolumn{2}{c}{InstructBLIP} \\
\cmidrule(lr){2-3} \cmidrule(lr){4-5} \cmidrule(lr){6-7}
& $\chairs$ $\downarrow$ & $\text{F}_{1}$ $\uparrow$ & $\chairs$ $\downarrow$ & $\text{F}_{1}$ $\uparrow$ & $\chairs$ $\downarrow$ & $\text{F}_{1}$ $\uparrow$ \\
\midrule
Baseline      & 41.8 & 77.5 & 31.8 & 69.9 & 61.4 & 71.6 \\
PAI           & 22.6 & 75.5 & 24.6 & 71.0 & 63.4 & 71.1 \\
OPERA         & 44.2 & 78.6 & 27.4 & 69.4 & 68.0 & 69.2 \\
VCD           & 56.0 & 71.1 & 31.0 & 70.0 & 61.6 & 71.6 \\
SPARC         & 45.6 & 78.9 & 26.8 & \textbf{72.5} & 70.0 & 70.0 \\
MiddleLayer   & \textbf{17.8} & 75.9 & 24.6 & 71.2 & 75.0 & 67.2 \\
Ours (w/o Ada.)  & 36.2 &\textbf{79.2} & \underline{23.0} & 71.3 & \underline{47.8} & \underline{74.1} \\
\rowcolor{gray!10}
Ours & \underline{18.2} & \textbf{79.2} &\textbf{17.0} &\underline{71.7} & \textbf{42.2} & \textbf{75.1} \\
\bottomrule
\end{tabular}}
\label{tab:chair_results}
\end{table*}

\begin{table}[t]
\centering
\caption{OPOPE performance of different models. Higher Accuracy($Acc$), Precision($Pre$), and F$_{\beta=0.2}$ indicate better object presence recognition and hallucination mitigation. The best results in each column are highlighted in \textbf{bold}, and the second-best are  \underline{underlined}.}
\renewcommand\arraystretch{1.15}
\setlength{\tabcolsep}{8pt}
\scalebox{0.5}{
\begin{tabular}{lccccccccc}
\toprule
\multirow{2}{*}{Method} & 
\multicolumn{3}{c}{LLaVA-1.5} & 
\multicolumn{3}{c}{MiniGPT-4} & 
\multicolumn{3}{c}{InstructBLIP} \\
\cmidrule(lr){2-4} \cmidrule(lr){5-7} \cmidrule(lr){8-10}
& $Acc$ $\uparrow$ & $Pre$ $\uparrow$ & F$_{\beta=0.2}$ $\uparrow$ 
& $Acc$ $\uparrow$ & $Pre$ $\uparrow$ & F$_{\beta=0.2}$ $\uparrow$
& $Acc$ $\uparrow$ & $Pre$ $\uparrow$ & F$_{\beta=0.2}$ $\uparrow$ \\
\midrule
Baseline & 76.75 & 86.17 & 85.03 & 67.94 & 87.43 & 83.94 & 73.94 & 83.26 & 82.06 \\
PAI      & 70.95 & 89.39 & 86.47 & 67.65 & 88.43 & 84.60 & 73.69 & 82.30 & 81.19 \\
OPERA    & \textbf{77.92} & 86.86 & 85.81 & 67.65 & 88.85 & 84.93 & 73.03 & 81.93 & 80.76 \\
VCD      & 73.57 & 83.98 & 82.59 & 67.86 & 87.39 & 83.88 & 74.23 & \underline{83.51} & 82.31 \\
SPARC    & \underline{77.04} & 86.13 & 85.04 & \underline{68.08} & 88.37 & 84.72 & \underline{74.57} & 82.87 & 81.84 \\
MiddleLayer   & 70.97 & \underline{89.43} & \underline{86.51} &  \underline{68.08} & \underline{89.61} & \underline{85.69} & \textbf{77.14} & 83.13 & \underline{82.44} \\
\rowcolor{gray!10}
Ours & 74.60 & \textbf{91.20} & \textbf{88.90} 
         & \textbf{68.13} & \textbf{90.74} & \textbf{86.59} 
         & 73.89 & \textbf{85.19} & \textbf{83.69} \\
\bottomrule
\end{tabular}}
\label{tab:three_models_comparison}
\end{table}
\subsection{Main Results on CHAIR}

Table~\ref{tab:chair_results} presents the quantitative results on the CHAIR benchmark across three representative LVLMs. 
Overall, our method strikes an effective balance between hallucination suppression and coverage of real objects. 
In particular, for LLaVA-1.5, the $\chairs$ decreases from 41.8 to 18.2, corresponding to a relative reduction of approximately 56\%. 
On MiniGPT-4 and InstructBLIP, our approach consistently achieves the lowest $\chairs$ values, confirming its robustness and effectiveness.

Moreover, across all models, our approach achieves the highest or second-highest $\text{F}_{1}$ values while significantly reducing hallucinations, showing that it effectively suppresses spurious objects without sacrificing faithful description quality.  
To further investigate this trade-off, we conduct detailed analyses on two aspects: the precision–recall trade-off and the $\chairs$–$\text{F}_{1}$ trade-off.

\begin{figure}[h!]
    \centering
    \includegraphics[width=0.4\textwidth]{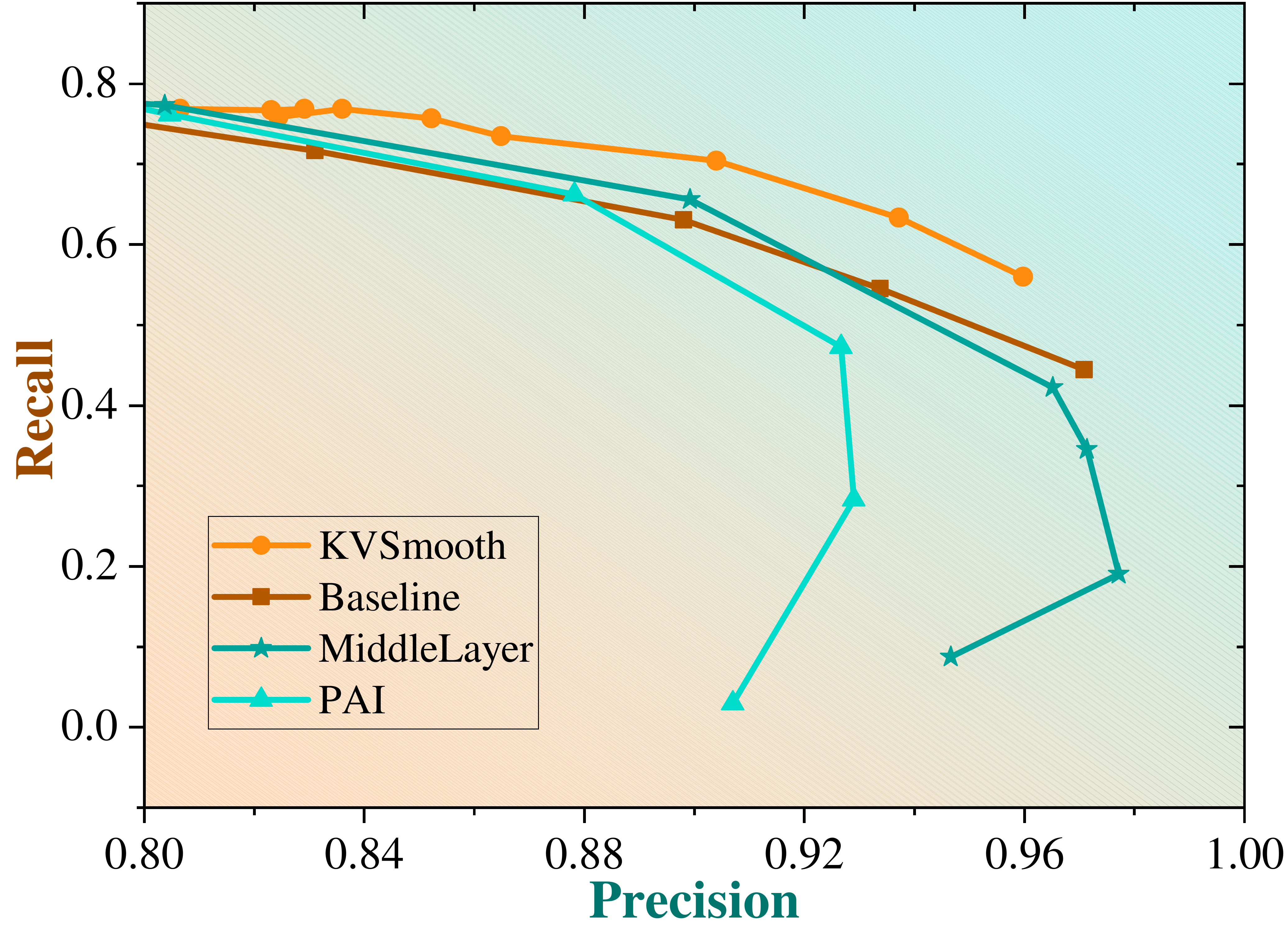}
    \caption{
    Precision–recall trade-off on LLaVA-1.5 (CHAIR benchmark). Curves nearer the top-right corner denote superior overall performance. \name attains a strong precision–recall balance, clearly surpassing all competing methods.
    }
    \label{figf1}
\end{figure}

\textbf{Analysis on Precision-Recall Trade-off.}
We further analyze model performance using Precision-Recall (PR) curves under the CHAIR benchmark (Figure~\ref{figf1}). Precision measures the model's ability to avoid hallucinations, while recall captures its ability to cover Ground Truth (GT) objects. We compare the baseline, two representative attention-reallocation methods, and our approach by sweeping their key parameters: the number of generated sentences (Baseline), the enhancement coefficient for visual tokens (PAI and MiddleLayer), and the EMA decay rate $\lambda_{\text{ref}}$ (\name). A larger area under the PR curve indicates a better balance between hallucination suppression and object coverage. Our method consistently achieves superior PR curves across all models, maintaining high precision with high recall. In contrast, while other methods can improve precision, they often result in significant recall degradation—sometimes even falling below the baseline at comparable precision levels.
\begin{figure*}[t]
    \centering
    \includegraphics[width=0.9\textwidth]{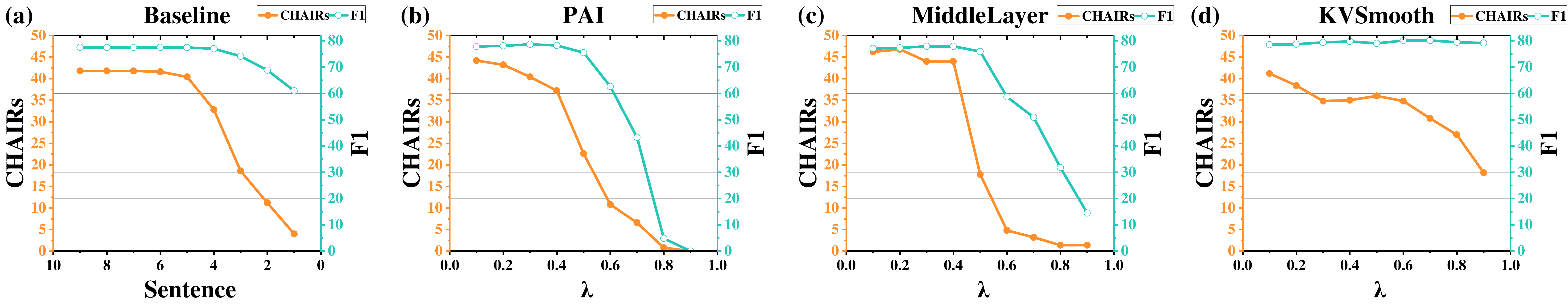}
    \caption{Sensitivity analysis of the hyperparameter $\lambda_{\text{ref}}$ for \name based on LLaVA-1.5 and comparisons of four methods in terms of the $\chairs$-${\text{F}_{1}}$ trade-off (CHAIR benchmark).
    It is evident that larger values of $\lambda_{\text{ref}}$ lead to stronger smoothing and improved hallucination mitigation. Moreover, our method consistently maintains the balance between precision and recall, demonstrating stability and reliability across different smoothing strengths.}
    \label{fig:lambda_sweep}
\end{figure*}

\textbf{Analysis on $\boldsymbol{\chairs}$-$\boldsymbol{\text{F}_{1}}$ Trade-off.}
We further explore the relationship between $\chairs$ and $\text{F}_{1}$ using the same parameter sweep settings as the precision–recall trade-off.  
Figure~\ref{fig:lambda_sweep} shows that other methods tend to suffer a significant drop in $\text{F}_{1}$ when reducing $\chairs$, whereas our approach can substantially reduce $\chairs$ while maintaining $\text{F}_{1}$ almost unchanged.  
This advantage stems from our adaptive mechanism (Section~\ref{sec:adaptive_coefficient}), which precisely identifies sink tokens that require smoothing and selectively mitigates hallucinations without affecting GT object descriptions.
In addition, Appendix~\ref{appendix:parameter_sensitivity} further reports the sensitivity analysis of the hyperparameter for \name, as well as the analysis of the $\chairs$–$\text{F}_{1}$ trade-off on the other two models. Besides, we provide additional qualitative generation examples in Appendix~\ref{appendix:qualitative_examples} and include an efficiency analysis in Appendix~\ref{appendix:xiaolv}, which shows that \name maintains inference speed and memory usage close to the baseline while incurring substantially lower overhead than competing methods.

\begin{table}[t]
\centering
\caption{
Amber and Object HalBench performance of different models.
Lower $\chair$ ($C$), $\chairs$ ($C_S$), $\chairsr$ ($C_{SR}$), $\chairi$ ($C_I$), $Hal$, and $Cog$ indicate better hallucination mitigation, while higher $Cover$ ($Cov$) and $CoCoNum$ ($Num$) indicate better object coverage.
The best results for each metric are highlighted in \textbf{bold}.
}
\label{tab:amber_objhal_results}
\renewcommand\arraystretch{1.15}
\setlength{\tabcolsep}{8pt}
\scalebox{0.5}{
\begin{tabular}{llcccccccc}
\toprule
Model & Method & 
\multicolumn{4}{c}{Amber} & 
\multicolumn{4}{c}{Object HalBench} \\
\cmidrule(lr){3-6} \cmidrule(lr){7-10}
& & $C\downarrow$ & $Cov\uparrow$ & $Hal\downarrow$ & $Cog\downarrow$ 
& $C_S\downarrow$ & $C_{SR}\downarrow$ & $C_I\downarrow$ & $Num\uparrow$ \\

\midrule

\multirow{2}{*}{LLaVA-1.5} 
& Baseline & 6.1 & 50.6 & 27.3 & 2.8 & 48.1 & 45.3 & 24.7 & 283 \\
& \multicolumn{1}{>{\columncolor{gray!10}}l}{Ours} 
& \multicolumn{1}{>{\columncolor{gray!10}}c}{\textbf{3.1}} 
& \multicolumn{1}{>{\columncolor{gray!10}}c}{\textbf{50.8}} 
& \multicolumn{1}{>{\columncolor{gray!10}}c}{\textbf{18.7}} 
& \multicolumn{1}{>{\columncolor{gray!10}}c}{\textbf{1.3}} 
& \multicolumn{1}{>{\columncolor{gray!10}}c}{\textbf{17.5}} 
& \multicolumn{1}{>{\columncolor{gray!10}}c}{\textbf{16.7}} 
& \multicolumn{1}{>{\columncolor{gray!10}}c}{\textbf{9.0}} 
& \multicolumn{1}{>{\columncolor{gray!10}}c}{\textbf{286}} \\
\midrule

\multirow{2}{*}{MiniGPT-4} 
& Baseline & \textbf{15.3} & \textbf{63.3} & 65.3 & 11.0 & 27.5 & 26.3 & 14.5 & \textbf{287} \\
& \multicolumn{1}{>{\columncolor{gray!10}}l}{Ours} 
& \multicolumn{1}{>{\columncolor{gray!10}}c}{17.8}
& \multicolumn{1}{>{\columncolor{gray!10}}c}{59.5} 
& \multicolumn{1}{>{\columncolor{gray!10}}c}{\textbf{39.6}} 
& \multicolumn{1}{>{\columncolor{gray!10}}c}{\textbf{5.1}} 
& \multicolumn{1}{>{\columncolor{gray!10}}c}{\textbf{15.8}} 
& \multicolumn{1}{>{\columncolor{gray!10}}c}{\textbf{15.7}} 
& \multicolumn{1}{>{\columncolor{gray!10}}c}{\textbf{8.6}} 
& \multicolumn{1}{>{\columncolor{gray!10}}c}{279} \\
\midrule

\multirow{2}{*}{InstructBLIP} 
& Baseline & 14.7 & \textbf{58.8} & 65.8 & 9.8 & 39.6 & 37.7 & 21.1 & \textbf{287} \\
& \multicolumn{1}{>{\columncolor{gray!10}}l}{Ours} 
& \multicolumn{1}{>{\columncolor{gray!10}}c}{\textbf{5.4}} 
& \multicolumn{1}{>{\columncolor{gray!10}}c}{57.6} 
& \multicolumn{1}{>{\columncolor{gray!10}}c}{\textbf{43.7}} 
& \multicolumn{1}{>{\columncolor{gray!10}}c}{\textbf{5.7}} 
& \multicolumn{1}{>{\columncolor{gray!10}}c}{\textbf{23.7}} 
& \multicolumn{1}{>{\columncolor{gray!10}}c}{\textbf{22.0}} 
& \multicolumn{1}{>{\columncolor{gray!10}}c}{\textbf{13.3}} 
& \multicolumn{1}{>{\columncolor{gray!10}}c}{278} \\
\bottomrule
\end{tabular}
}
\end{table}


\begin{table}[t]
\centering
\caption{Performance comparison under different smoothing methods. 
Lower $\chairs$ ($C_s$) indicates better hallucination mitigation, and higher $\text{F}_{1}$ indicates better overall performance.  
The best results in each column are highlighted in \textbf{bold}, and the second-best are \underline{underlined}.}
\label{tab:method}
\renewcommand\arraystretch{1.15}
\setlength{\tabcolsep}{8pt}
\scalebox{0.75}{
\begin{tabular}{lccccccccc}
\toprule
\multirow{2}{*}{Method} 
& \multicolumn{2}{c}{LLaVA-1.5} 
& \multicolumn{2}{c}{MiniGPT-4} 
& \multicolumn{2}{c}{InstructBLIP} \\
\cmidrule(lr){2-3} \cmidrule(lr){4-5} \cmidrule(lr){6-7}
 & $C_s$↓ & $\text{F}_{1}$↑ 
 & $C_s$↓ & $\text{F}_{1}$↑ 
 & $C_s$↓ & $\text{F}_{1}$↑ \\
\midrule
Attention output ($o_t$)   & \underline{33.8} & 74.7 
                          & \underline{19.8} & 66.5 
                          & 84.4 & 61.2 \\
Key ($K_t$)            & 35.6 & \textbf{79.4} 
                          & 24.0 & \underline{70.6} 
                          & \underline{73.2} & \underline{70.2} \\
\rowcolor{gray!10}
Key-Value ($K_t, V_t$)   & \textbf{18.2} & \underline{79.2} 
                          & \textbf{17.0} & \textbf{71.7} 
                          & \textbf{42.2} & \textbf{75.1} \\
\bottomrule
\end{tabular}
}
\end{table}

\subsection{Generalization Experiments}
To further verify the generalization ability of our approach beyond the CHAIR benchmark, we extend the evaluation to three complementary hallucination benchmarks—\textbf{OPOPE}, \textbf{AMBER}, and \textbf{Object HalBench}—which jointly assess object presence verification, multi-scenario hallucination robustness, and GPT-assisted object grounding.

\textbf{OPOPE.}  
As shown in Table~\ref{tab:three_models_comparison}, our method achieves the highest average Accuracy and F$_{\beta=0.2}$ across random, popular, and adversarial object sets,  demonstrating strong hallucination mitigation in all three scenarios.

\textbf{AMBER.}  
As shown in Table~\ref{tab:amber_objhal_results}, it can be observed that our method substantially reduces hallucinated objects while retaining coverage of GT objects. 
Specifically, for LLaVA-1.5, our approach not only yields a remarkable reduction in $\chair$ and $Hal$ (fewer hallucinations) but also slightly increases $Cover$, with more GT objects correctly described while suppressing hallucinations.
For MiniGPT-4 and InstructBLIP, our method also consistently lowers hallucination metrics compared to the baseline while maintaining competitive coverage of real objects.

\textbf{Object HalBench.}  
As shown in Table~\ref{tab:amber_objhal_results}, \name significantly decreases the sentence-level hallucination rate $\chairsr$ by 63.1\%, 40.3\%, and 41.6\% for LLaVA-1.5, MiniGPT-4, and InstructBLIP, respectively. These consistent improvements across all metrics confirm the effectiveness and generalizability of our approach in mitigating multi-modal hallucinations under varied evaluation protocols.

\subsection{Ablation Study}
\subsubsection{Effect of Smoothing Method}
\label{subsec: effect of smoothing method}

To determine the optimal positions for applying EMA updates, we investigate the impact of updating different internal states on hallucination mitigation. Specifically, EMA is applied separately to the key $K_t$, the attention output $o_t$, and jointly to both key and value ($K_t, V_t$). Their performance is evaluated in terms of $\chairs$ and $\text{F}_{1}$ scores on the CHAIR benchmark, as summarized in Table~\ref{tab:method}.

From the results, it can be seen that applying EMA solely to the key vectors is less effective than updating both key and value vectors simultaneously. In contrast, applying EMA directly to the hidden states leads to a severe drop in recall, indicating a significant reduction in the model's ability to retain correct object information. These findings suggest that joint EMA updates on the key and value vectors strike the best balance, effectively mitigating hallucinations while preserving faithful object descriptions.

\subsubsection{Module Ablation}
Our method comprises two main components: (1) EMA smoothing on KV-Cache (Section~\ref{sec:ema}) and (2) entropy-guided coefficient adaptation (Section~\ref{sec:adaptive_coefficient}).

To evaluate the contribution of the adaptive design, we set the EMA update coefficient $\lambda$ as a constant and select the value that yields the highest $\text{F}_{1}$ score for comparison. As shown in Table~\ref{tab:chair_results}, even with the best constant $\lambda$, our adaptive mechanism still outperforms it. This demonstrates that the adaptive design can precisely identify tokens more likely to induce hallucinations and apply stronger smoothing to them. In contrast, a fixed $\lambda$ may over-smooth genuine object regions or under-smooth hallucinated ones, thereby limiting the model’s ability to generate accurate and detailed descriptions.
\section{Conclusion}
\label{sec:conclusion}

This work investigated the critical issue of hallucination in multi-modal models, with a specific focus on semantic drift during long text generation. We established three pivotal observations: \textbf{Obs1} reveals that both the mean and variance of logit scores for hallucinated objects progressively increase during generation; \textbf{Obs2} identifies a strong correlation between attention row-entropy and attention aggregation patterns; and \textbf{Obs3} demonstrates a causal coupling between attention row-entropy and hallucination amplification. Leveraging these insights, we proposed \name, a lightweight, training-free, and plug-and-play approach that dynamically computes row-entropy during generation and applies adaptive EMA smoothing to the Key and Value vectors. Extensive experiments validate that \name not only achieves state-of-the-art performance in mitigating hallucinations but also preserves comprehensive and accurate coverage of real objects.




{
    \small
    \bibliographystyle{ieeenat_fullname}
    \bibliography{main}
}
\clearpage
\setcounter{page}{1}
\maketitlesupplementary
\appendix

\renewcommand{\thesection}{\Alph{section}}  
\setcounter{section}{0} 

\renewcommand{\thesubsection}{\thesection.\arabic{subsection}}

\section{The Detailed Derivation of EMA Smoothing on Hidden States}
\label{appendix:detail derivation of ema smoothing}
During inference, this growth manifests as increasingly large token-wise logit moments: the model state oscillates violently, drifts away from the visual prior, and deviates from the true image information through sudden jumps.
To keep the decoding trajectory ideally smooth, we formalize the following hypothesis:  
\begin{equation}
    \label{eq:sup_gauss}
    h_t=h_{t-1} + \epsilon_t, \epsilon_t\sim N(0, \sigma^2_p).
\end{equation}
Consequently, the transition prior is  
\begin{equation}
\label{eq:sup_priori}
    \begin{split}
    P(h_t | h_{t-1})=N(h_t;h_{t-1},\sigma^2_p)\propto \\
    \exp\left(-\frac{1}{2\sigma^2_p}||h_t - h_{t-1}||^2\right). 
    \end{split}
\end{equation}
Under this prior, the maximum-a-posteriori (MAP) estimate of $h_t$ is given by  
\begin{equation}
\label{eq:sup_map}
\hat{h_t}=\argmax_{h_t}\left[ \underbrace{\log P(o_t|h_t)}_{\text{likelihood}} + \underbrace{\log P(h_t|h_{t-1})}_{\text{smoothed prior}} \right],
\end{equation}
where $o_t$ denotes the raw hidden state observed at step $t$.  
Assuming both terms are Gaussian, we have  
\begin{equation}
    \label{eq:sup_observe}
    P(o_t|h_t) \propto \exp(-\frac{1}{2\sigma^2_o}||o_t - h_{t}||^2),
\end{equation}
\begin{equation}
    \label{eq:sup_smoothed priori}
    P(h_t|h_{t-1}) \propto \exp(-\frac{1}{2\sigma^2_p}||h_t - h_{t-1}||^2).
\end{equation}
Substituting \eqref{eq:sup_observe} and \eqref{eq:sup_smoothed priori} into \eqref{eq:sup_map} yields  
\begin{equation}
    \label{eq:sup_insert map}
    \hat{h_t} = \argmax_{h_t}\left[ 
    -\frac{1}{2\sigma^2_o}||o_t - h_{t}||^2 -\frac{1}{2\sigma^2_p}||h_t - h_{t-1}||^2
    \right].
\end{equation}
Taking the derivative of \eqref{eq:sup_insert map} with respect to $h_t$ and setting it to zero, we obtain:
\begin{equation}
    \label{eq:sup_partial derivative}
    \begin{split}
    \frac{\partial}{\partial h_t}\left[ 
    -\frac{1}{2\sigma^2_o}||o_t - h_{t}||^2 -\frac{1}{2\sigma^2_p}||h_t - h_{t-1}||^2
    \right] = 0 \\
    \frac{1}{\sigma^2_o}(o_t - h_t) - \frac{1}{\sigma^2_p}(h_t - h_{t-1}) = 0.
    \end{split}
\end{equation}
Solving for $\hat{h_t}$, we obtain  
\begin{equation}
    \label{eq:sup_preema}
    \hat{h_t} = \frac{\sigma^2_p}{\sigma^2_p + \sigma^2_o}o_t + \frac{\sigma^2_o}{\sigma^2_p + \sigma^2_o}h_{t-1}.
\end{equation}
Defining  
\begin{equation}
    \lambda_t = \frac{\sigma^2_o}{\sigma^2_p + \sigma^2_o},  
\end{equation}
the estimate reduces to the exponential moving average (EMA) form  
\begin{equation}
    \label{eq:sup_ema}
    \hat{h_t} = (1 - \lambda_t)o_t + \lambda_t h_{t - 1}.
\end{equation}
Thus, when the likelihood is Gaussian in $h_t$, the MAP estimator coincides exactly with an exponential moving average.


\section{Benchmark Details}
\label{appendix:benchmark_details}

We briefly describe the four benchmarks used for evaluation:
\paragraph{CHAIR.}
CHAIR~\cite{chair} is a classic benchmark for measuring \textbf{object hallucination} in image captioning tasks.  
It defines two key metrics: 
\begin{itemize}
    \item $\boldsymbol{\chairs}$ (Sentence-level Hallucination) — the proportion of captions that contain at least one hallucinated object.
    \item $\boldsymbol{\text{F}_{1}}$ — a balanced measure reflecting both the accuracy and completeness of generated captions. In CHAIR, \textit{precision} quantifies the proportion of generated objects that correctly appear in the ground-truth annotations, while \textit{recall} measures how many ground-truth objects are successfully mentioned in the generated captions. The ${\text{F}_{1}}$ score captures the overall trade-off between these two factors.
\end{itemize}
In this work, following \cite{opera}, we randomly sample 500 images from the COCO 2014 validation set and prompt the LVLMs with “Please describe the image in detail.” The maximum output length is set to 512 tokens.

\paragraph{OPOPE.}
OPOPE~\cite{opope} extends POPE \cite{pope} by transforming its interactive yes/no polling mechanism into an \textbf{offline evaluation}.  
It retains POPE’s three sampling strategies — \textit{random}, \textit{popular}, and \textit{adversarial} — but checks whether sampled positive and negative objects appear in the model-generated image descriptions instead of interacting with the model.  
Following~\cite{opope}, we report the following metrics:
\begin{itemize}
    \item \textbf{Accuracy} — overall correctness of object presence identification.
    \item \textbf{Precision} — proportion of correctly identified positive objects among all predicted positives.
    \item \textbf{F$_{\beta=0.2}$}— defined as 
    \[
    F_{\beta} = (1+\beta^2)\frac{\text{Precision}\cdot \text{Recall}}{\beta^2 \cdot \text{Precision} + \text{Recall}}, \quad \beta=0.2,
    \] 
    which reduces the impact of false negatives following~\cite{opope}.
\end{itemize}
All reported numbers are averaged over the three sampling methods (random, popular, and adversarial).

\paragraph{AMBER.} 
AMBER~\cite{amber} is a benchmark for evaluating hallucinations in vision-language models from both generative and discriminative perspectives. In this work, we focus on the generative task setting to assess model performance.  
The dataset encompasses 14 major object categories, featuring a balanced distribution that mitigates the significant long-tail issue. Compared to existing benchmarks, AMBER extends coverage to categories such as Nature, Architecture, and Street View, and provides richer annotations in others—for example, the Fruit category includes over a dozen common fruits, whereas prior datasets only cover three types.  

Following \cite{amber}, we adopt four metrics for evaluation:  
\begin{itemize}
    \item \textbf{CHAIR} — object hallucination rate in responses.
    \item \textbf{Cover} — proportion of objects mentioned in the response relative to annotated objects, reflecting faithful object coverage.
    \item \textbf{Hal} — proportion of responses containing hallucinations.
    \item \textbf{Cog} — measures the extent to which hallucinated objects align with common human cognitive biases.
\end{itemize}

\paragraph{Object HalBench.}  
Object HalBench (ObjHal)~\cite{chair} is a diverse benchmark for evaluating \textbf{object hallucination robustness} under different prompt styles. It assesses models on 300 image–text pairs using eight varied prompts, providing a stable and comprehensive evaluation.  
Here, GPT-4~\cite{gpt4} is used to extract visible objects from the captions generated by the model.  

The benchmark reports four main metrics:  
\begin{itemize}
    \item $\boldsymbol{\chairs}$ — Image-level hallucination rate: proportion of images containing at least one hallucinated object.  
    \item $\boldsymbol{\chairsr}$ — Similar to $\chairs$ but excludes sentences without any object words; measures hallucination among sentences containing at least one MSCOCO object word.  
    \item $\boldsymbol{\chairi}$ — Proportion of hallucinated objects among all generated object words.  
    \item \textbf{CoCoNum} — Number of captions containing at least one COCO object.  
\end{itemize}

\label{appendix:xiaolv}
\section{Efficiency Comparisons}
To further assess the computational efficiency and resource usage of our approach, we compare its average per-caption inference time, token-level latency, throughput, and memory consumption with several representative hallucination mitigation methods, as summarized in Table~\ref{tab:dataset}.

As illustrated in the table, our method achieves a favorable trade-off between performance and efficiency. It maintains a comparable runtime and memory footprint to the baseline model while significantly reducing hallucinations. Notably, compared with more complex attention redistribution approaches such as SPARC and PAI, our approach requires substantially less computation time and memory overhead. This demonstrates that our smoothing mechanism is lightweight and can be seamlessly integrated into existing vision-language models without compromising efficiency.

\begin{table}[t]
\centering
\caption{Comparison of token generation efficiency and resource usage across different methods on LLaVA-1.5 (CHAIR benchmark). 
Our method delivers faster inference and lower memory cost than other training-free hallucination mitigation methods.}
\label{tab:dataset}
\renewcommand\arraystretch{1.1}
\scalebox{0.85}{
\begin{tabular}{l r r r r}
\toprule
Method & Avg. Time & Peak Memory & Latency & Throughput \\
       & (s/caption) & (MB) & (ms/token) & (token/s) \\
\midrule
Baseline & \textbf{3.36}&\underline{14629.21}   &  \textbf{31.24}     &  \textbf{32.06} \\
PAI      &  6.68       & 15003.25    &  56.44     & 17.73 \\
OPERA    & 34.62      & 19531.97    & 313.78     &  3.43 \\
SPARC    &  4.13      & 14644.61    &  35.20     & 28.44 \\
\rowcolor{gray!10}
Ours     &  \underline{3.61}      & \textbf{14625.06}    &  \underline{34.33}     & \underline{29.18} \\
\bottomrule
\end{tabular}}
\end{table}








\begin{figure}[h] 
    \centering
    \includegraphics[width=0.5\textwidth]{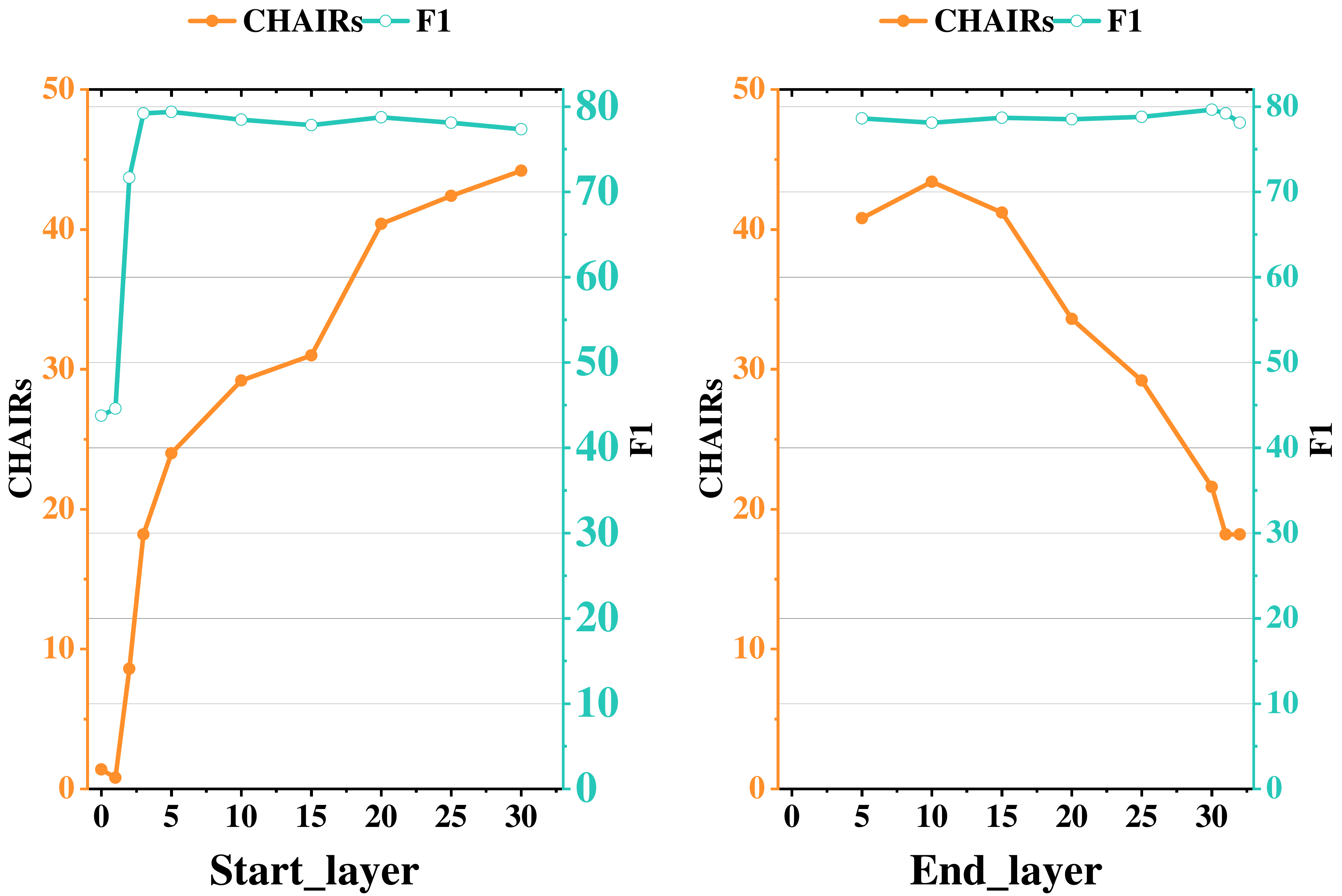} 
    \caption{Sensitivity analysis of the layer range for \name on LLaVA-1.5  (CHAIR benchmark). Left: sensitivity to start layer ($L_{\text{start}}$); Right: sensitivity to end layer ($L_{\text{end}}$).} 
    \label{fig:layer_sensitivity} 
\end{figure}

\begin{figure*}[t]
    \centering
    \includegraphics[width=0.85\textwidth]{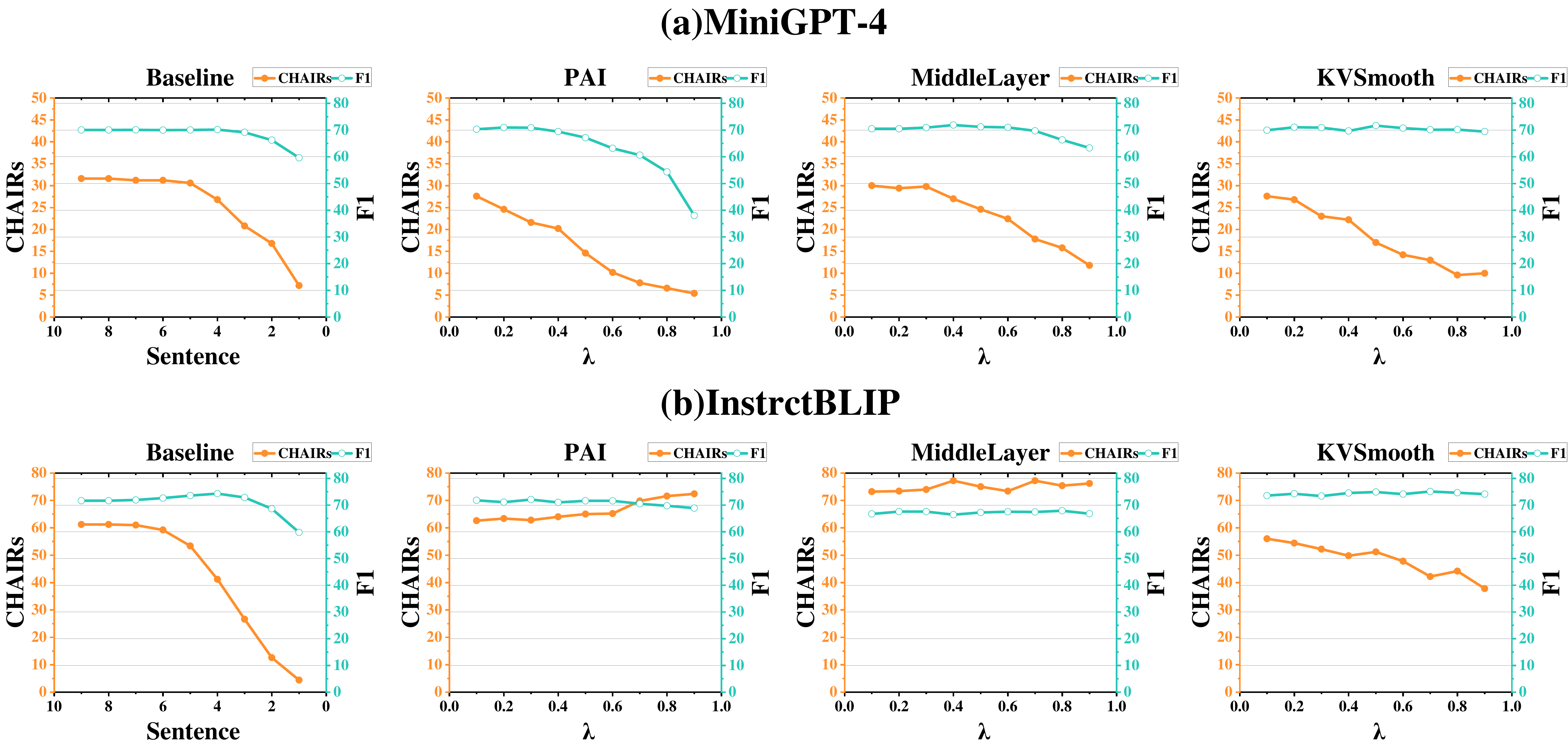}
    \caption{
    Sensitivity analysis of the hyperparameter $\lambda_{\text{ref}}$ for \name based on MiniGPT-4 (a) and InstructBLIP (b) and comparisons of four methods in terms of the $\chairs$–$\text{F}_{1}$ trade-off (CHAIR benchmark). 
    Larger values of $\lambda_{\text{ref}}$ lead to stronger smoothing and more effective hallucination mitigation. Moreover, our method consistently maintains a favorable balance between precision and recall: it reduces hallucinations while preserving a high $\text{F}_{1}$ score, demonstrating diverse and reliable object coverage across different smoothing strengths.
    }
    \label{fig:lambda_sweep_app}
\end{figure*}

\label{appendix:parameter_sensitivity}
\section{Parameter Sensitivity}
To evaluate the robustness of our approach, we conduct a sensitivity analysis on key hyperparameters across three MLLMs, including the EMA decay rate $\lambda_{\text{ref}}$, which determines the strength of the smoothing effect, and the start layer ($L_{\text{start}}$) and end layer ($L_{\text{end}}$), which define the range of layers where EMA is applied. We vary each parameter while keeping the others fixed to examine its influence on overall performance. 
\subsection{Effect of the Reference Smoothing Coefficient $\lambda_{\text{ref}}$}
\label{appendix:ref_sensitivity}
Specifically, we analyze the influence of the decay threshold range by varying $\lambda_{\text{ref}}$ and evaluating \name on the CHAIR benchmark, measuring both $\chairs$ and $\text{F}_{1}$ metrics. As shown in Figure~\ref{fig:lambda_sweep} and Figure~\ref{fig:lambda_sweep_app}, the $\chairs$ score consistently decreases as $\lambda_{\text{ref}}$ increases, while the $\text{F}_{1}$ score remains nearly unchanged. This indicates that a larger $\lambda_{\text{ref}}$ leads to stronger smoothing effects and more effective hallucination mitigation. Moreover, despite enhanced hallucination suppression, \name maintains a stable balance between precision and recall, demonstrating robustness across different smoothing strengths.
\subsection{Effect of Start and End Layers($L_{\text{start}}$, $L_{\text{end}}$).}
\label{appendix:layer_sensitivity}
To determine the optimal layer range for applying EMA updates, we conduct a two-dimensional layer sweep. 
First, we fix the end layer $L_{\text{end}}$ at 31 and vary the start layer $L_{\text{start}}$ from 0 to 30. 
Then, we fix $L_{\text{start}}$ at 3 and vary $L_{\text{end}}$ from 5 to 32. 
As shown in Figure~\ref{fig:layer_sensitivity}, $\chairs$ decreases gradually as more layers are included in the EMA updates, demonstrating stronger hallucination suppression. 
However, applying EMA at the very early (0–2) or final (32) layers causes a noticeable drop in $\text{F}_{1}$,  indicating that excessive smoothing in these regions impairs the model’s ability to represent real objects and degrades caption quality.
%
\section{The Algorithm of \name}
\label{appendix:fake}

In this section, we present the full algorithmic details of \name. The algorithm consists of two components:  
\textbf{Algorithm~1: EMA Smoothing on KV-Cache}, corresponding to Section~\ref{sec:final} (\emph{The Final \name Method}); and  
\textbf{Algorithm~2: Attention Forward with Adaptive EMA Smoothing}, corresponding to Section~\ref{sec:fomulation of mmlms generation} (\emph{Formulation of MLLMs Generation}) and Section~\ref{sec:adaptive_coefficient} (\emph{Entropy-Guided Coefficient Adaptation}).

\begin{algorithm}[h]
\small
\caption{EMA Smoothing on KV-Cache}
\label{alg:ema_kv}
\KwIn{KV-cache $\mathcal{C}_{t}^l = \bigl(K_{1:t}^{(l,h)}, V_{1:t}^{(l,h)}\bigr)_{h=1}^H$, smoothing coefficient $\tilde{\lambda}_{t}^l$}
\KwOut{Smoothed KV-cache $\mathcal{C}_t^l = \bigl(K_{1:t}^{(l,h)}, V_{1:t}^{(l,h)}\bigr)_{h=1}^H$}

\BlankLine
\textbf{1. EMA update for each head $h$:}
\[
\hat{K}_{t}^{(l,h)} \leftarrow 
(1-\tilde{\lambda}_{t}^l)\,K_{t}^{(l,h)}
+ \tilde{\lambda}_{t}^l\, K_{t-1}^{(l,h)},
\]
\[
\hat{V}_{t}^{(l,h)} \leftarrow 
(1-\tilde{\lambda}_{t}^l)\,V_{t}^{(l,h)}
+ \tilde{\lambda}_{t}^l\, V_{t-1}^{(l,h)}.
\]

\BlankLine
\textbf{2. Update smoothed cache:}
\[
K_{t}^{(l,h)} \leftarrow \hat{K}_{t}^{(l,h)},\qquad
V_{t}^{(l,h)} \leftarrow \hat{V}_{t}^{(l,h)}.
\]

\BlankLine
\Return{$\mathcal{C}_t^l$}

\end{algorithm}

\begin{algorithm*}[h]
\small
\caption{Attention Forward with Adaptive EMA Smoothing}
\label{alg:adaptive_ema}
\KwIn{Hidden state $h_t$, mask $M$, KV-cache $\mathcal{C}_{t-1}^l$, entropy queue $S^l$, reference coefficient $\lambda_{\text{ref}}$}
\KwOut{Updated hidden state $h_t^l$, attention weights $\alpha_t^{(l,h)}$, KV-cache $\mathcal{C}_t^l$, entropy queue $S^l$}

\BlankLine
\textbf{1. Compute Q, K, V for each head $h$:}
\[
Q_t^{(l,h)} = h_t W_Q^{(l,h)},\quad
K_t^{(l,h)} = h_t W_K^{(l,h)},\quad
V_t^{(l,h)} = h_t W_V^{(l,h)}.
\]

\BlankLine
\textbf{2. Update KV-cache:} Append $(K_t^{(l,h)}, V_t^{(l,h)})$ to $\mathcal{C}_{t-1}^l$ to obtain $\mathcal{C}_t^l$.

\BlankLine
\textbf{3. Compute attention scores:}
\[
A_{t,j}^{(l,h)} = 
\frac{\bigl(Q_t^{(l,h)}\bigr)^\top K_j^{(l,h)}}{\sqrt{d}} + M_{t,j}.
\]

\BlankLine
\textbf{4. Compute attention weights and row-entropy:}
\[
\alpha_{t,j}^{(l,h)}
= \mathrm{Softmax}_j\!\bigl(A_{t,j}^{(l,h)}\bigr),
\]
\[
z_{t,j}^{\,l}
= -\frac{1}{H}
  \sum_{h=0}^{H-1}
  \sum_{i=0}^{L-1}
  \alpha_{i,j}^{(l,h)}
  \log\!\bigl(\alpha_{i,j}^{(l,h)} + \varepsilon\bigr).
\]
Insert $z_{t,j}^l$ into the entropy queue $S^l$.

\BlankLine
\textbf{5. Determine adaptive smoothing coefficient:}
Compute the percentile rank $k$ of $z_{t,j}^l$ in $S^l$.
\[
\hat{\lambda}_t^l = k / |S^l|,
\]
\[
\tilde{\lambda}_t^l = \mathrm{clip}
\bigl(\hat{\lambda}_t^l,\ 
\lambda_{\text{ref}} - 0.2,\ 
\lambda_{\text{ref}} + 0.2 \bigr).
\]

\BlankLine
\textbf{6. Compute attention output:}
\[
o_t^{(l,h)} 
= \sum_{j=0}^{L-1} \alpha_{t,j}^{(l,h)} V_j^{(l,h)},
\]
\[
h_t^l = \mathrm{Concat}_h(o_t^{(l,h)})\, W_O^l.
\]

\BlankLine
\textbf{7. Apply EMA smoothing for each head:}
\[
(K_{1:t}^{(l,h)}, V_{1:t}^{(l,h)})_{h=1}^H
\leftarrow
\mathrm{EMA Smoothing on}
\bigl((K_{1:t}^{(l,h)}, V_{1:t}^{(l,h)})_{h=1}^H, \tilde{\lambda}_t^l\bigr).
\]

\BlankLine
\Return{$h_t^l,\ \alpha_t^{(l,h)},\ \mathcal{C}_t^l,\ S^l$}

\end{algorithm*}

\section{Qualitative Results}
To illustrate the effectiveness of \name, we show representative examples where \textcolor{myorange}{orange} denotes baseline hallucinations and \textcolor{mycyan}{cyan} denotes the corrections by \name, demonstrating its ability to fix factual errors while preserving coherence.
\label{appendix:qualitative_examples}
\begin{figure*}[t]
\centering
\includegraphics[height=0.31\textheight]{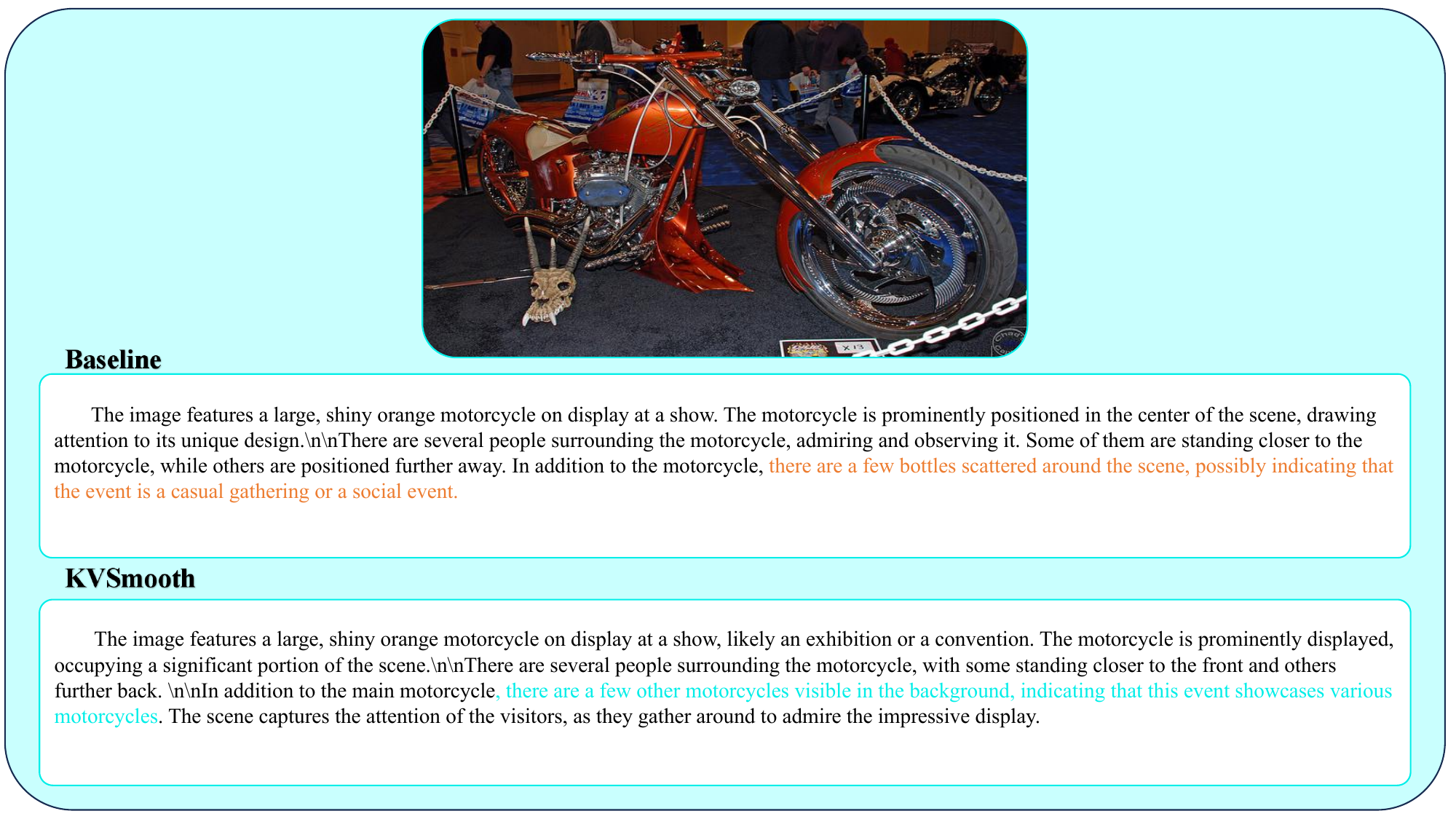}
\includegraphics[height=0.31\textheight]{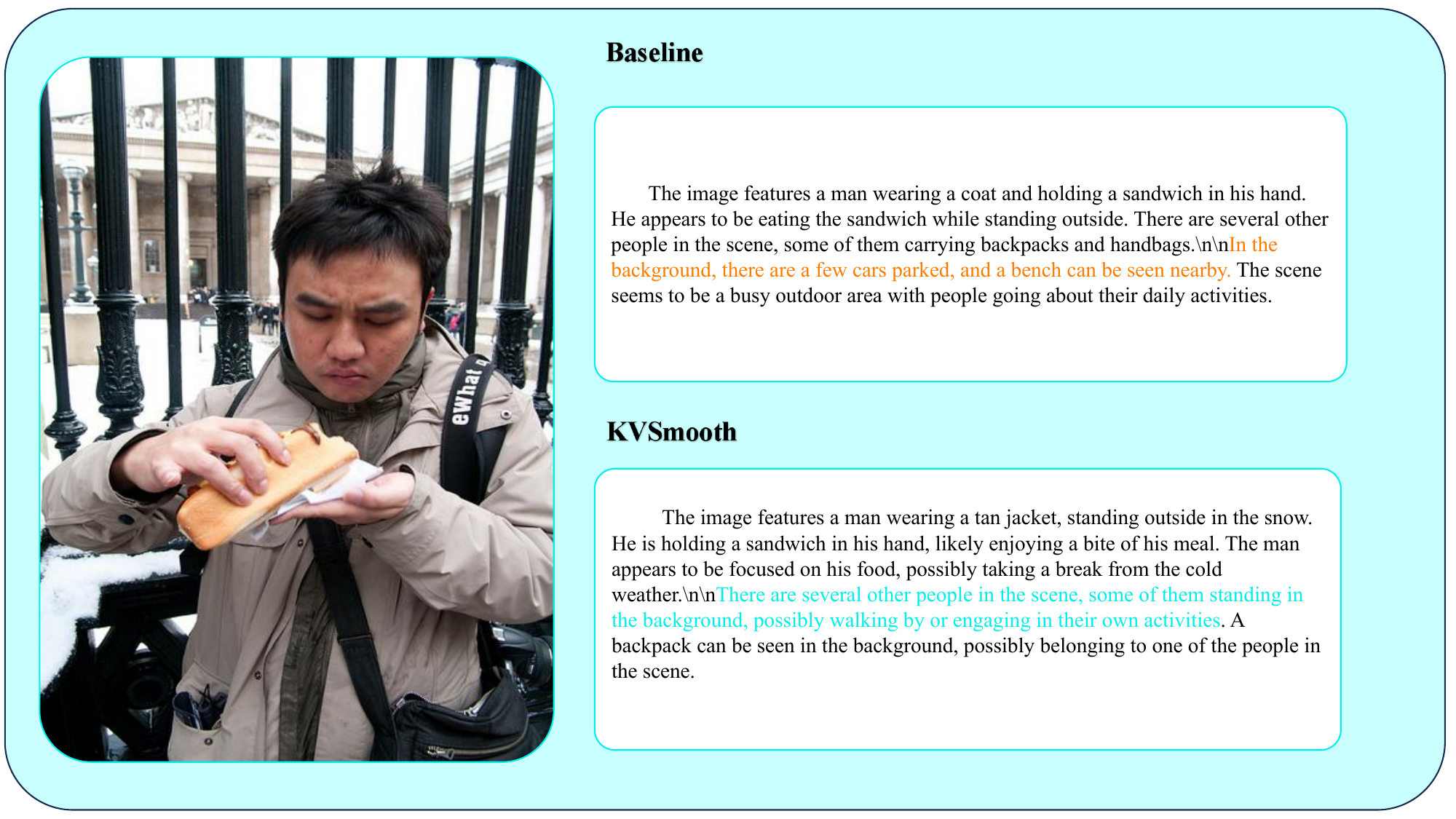}
\includegraphics[height=0.31\textheight]{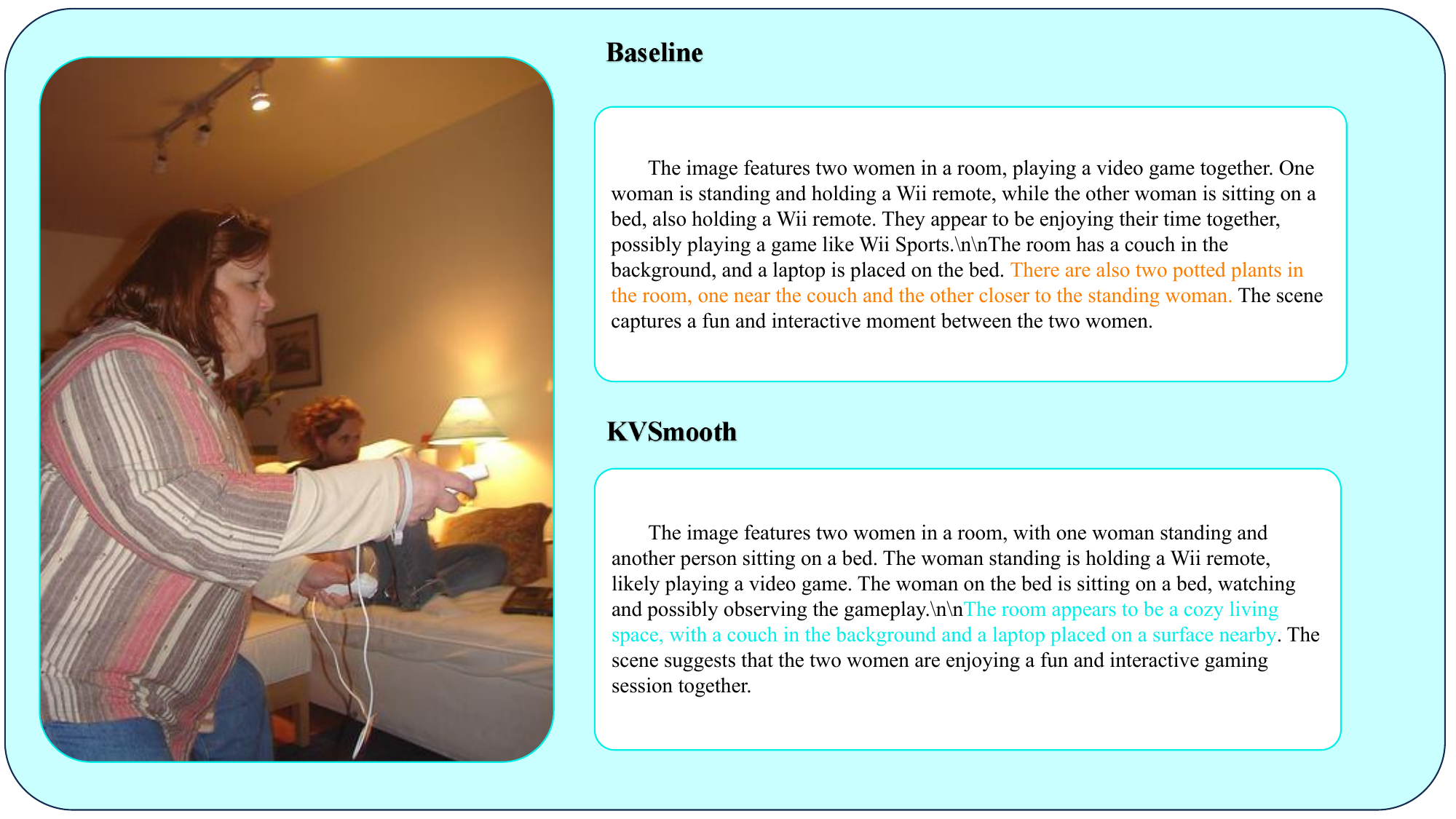}
\caption{Qualitative results on LLaVA-1.5: baseline (w/o \name) vs.\ \name.}
\label{fig:qualitative_llava-1.5}
\end{figure*}
\begin{figure*}[t]
\centering

\includegraphics[height=0.31\textheight]{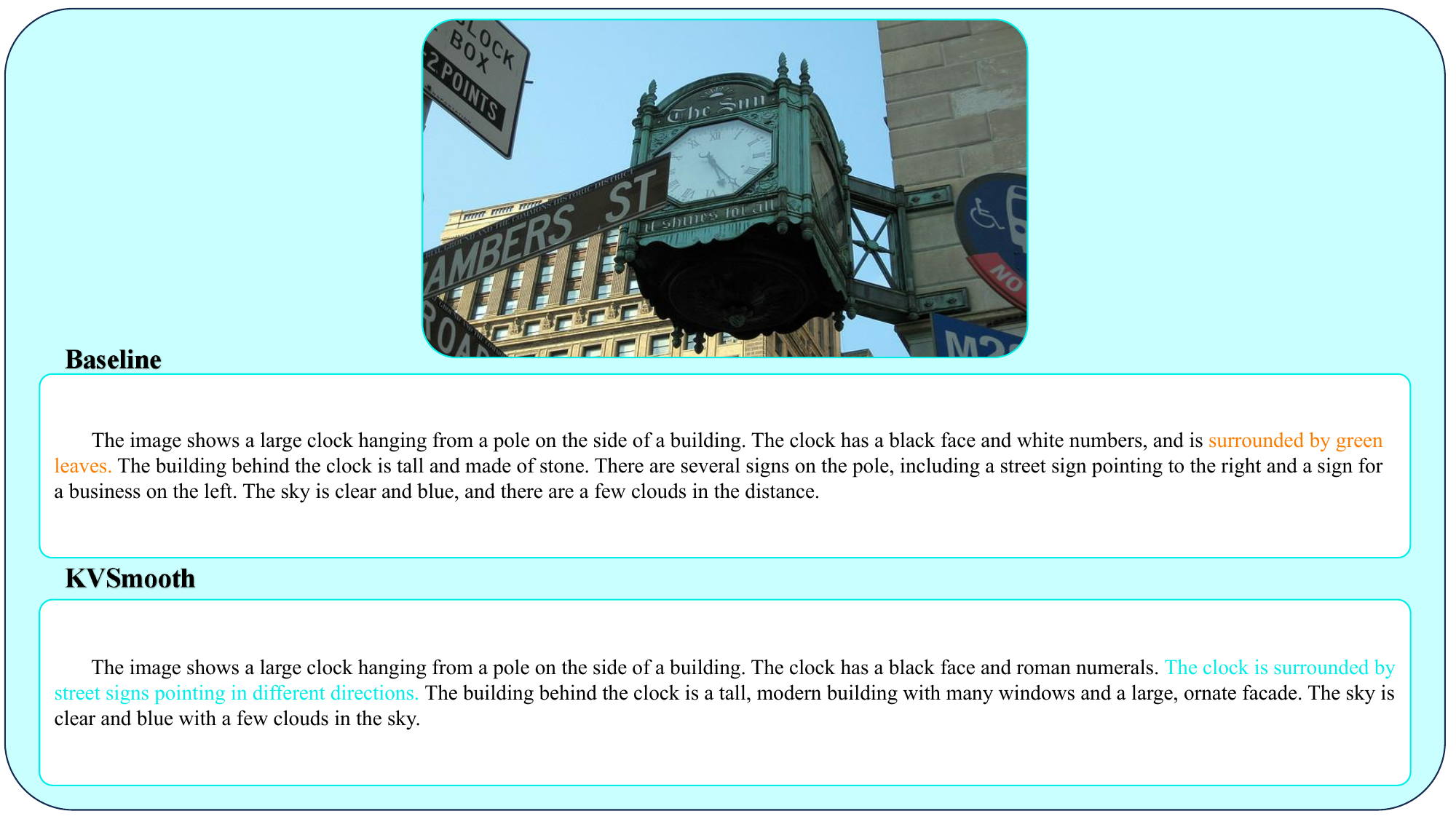}
\includegraphics[height=0.31\textheight]{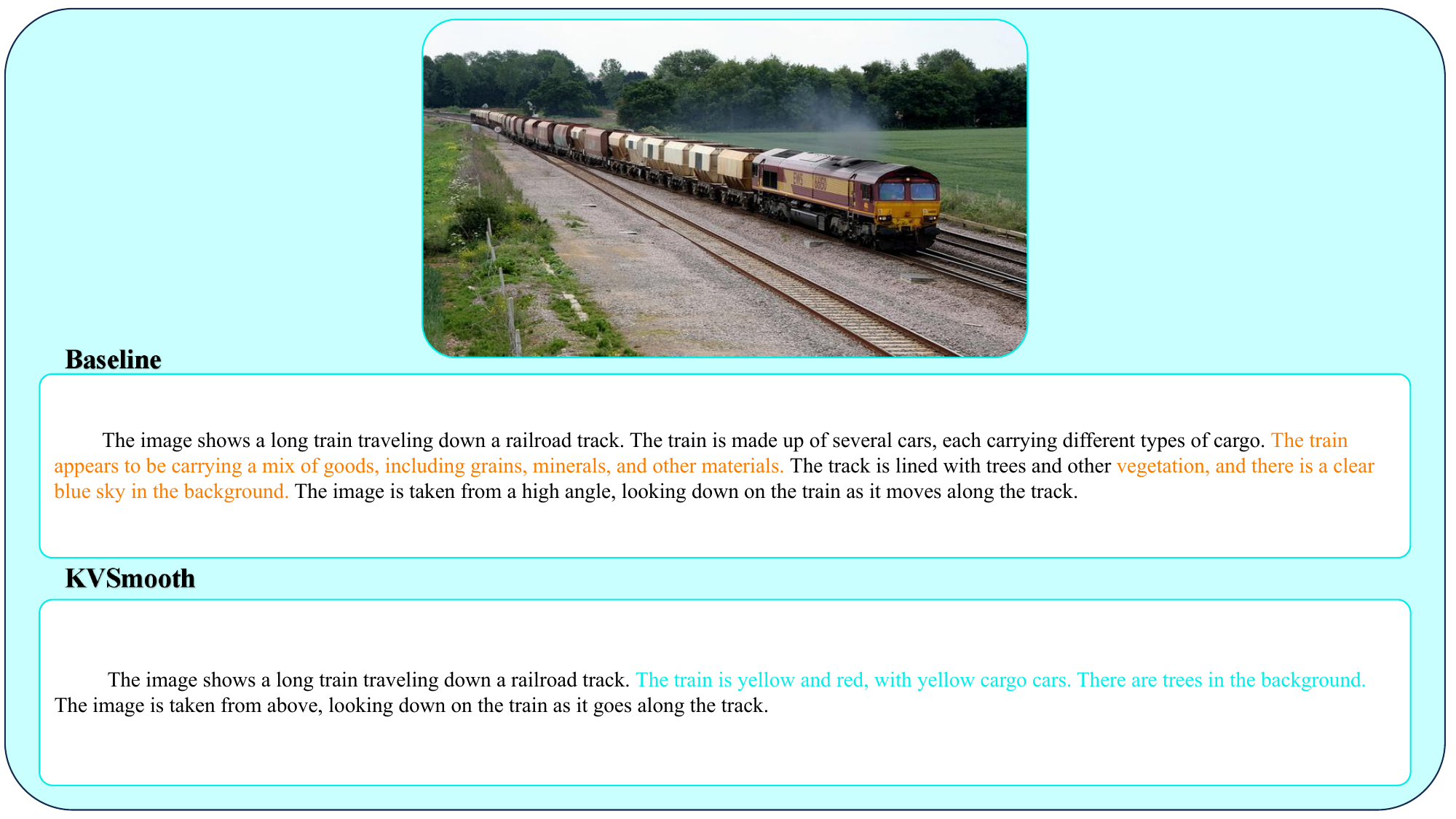}
\includegraphics[height=0.31\textheight]{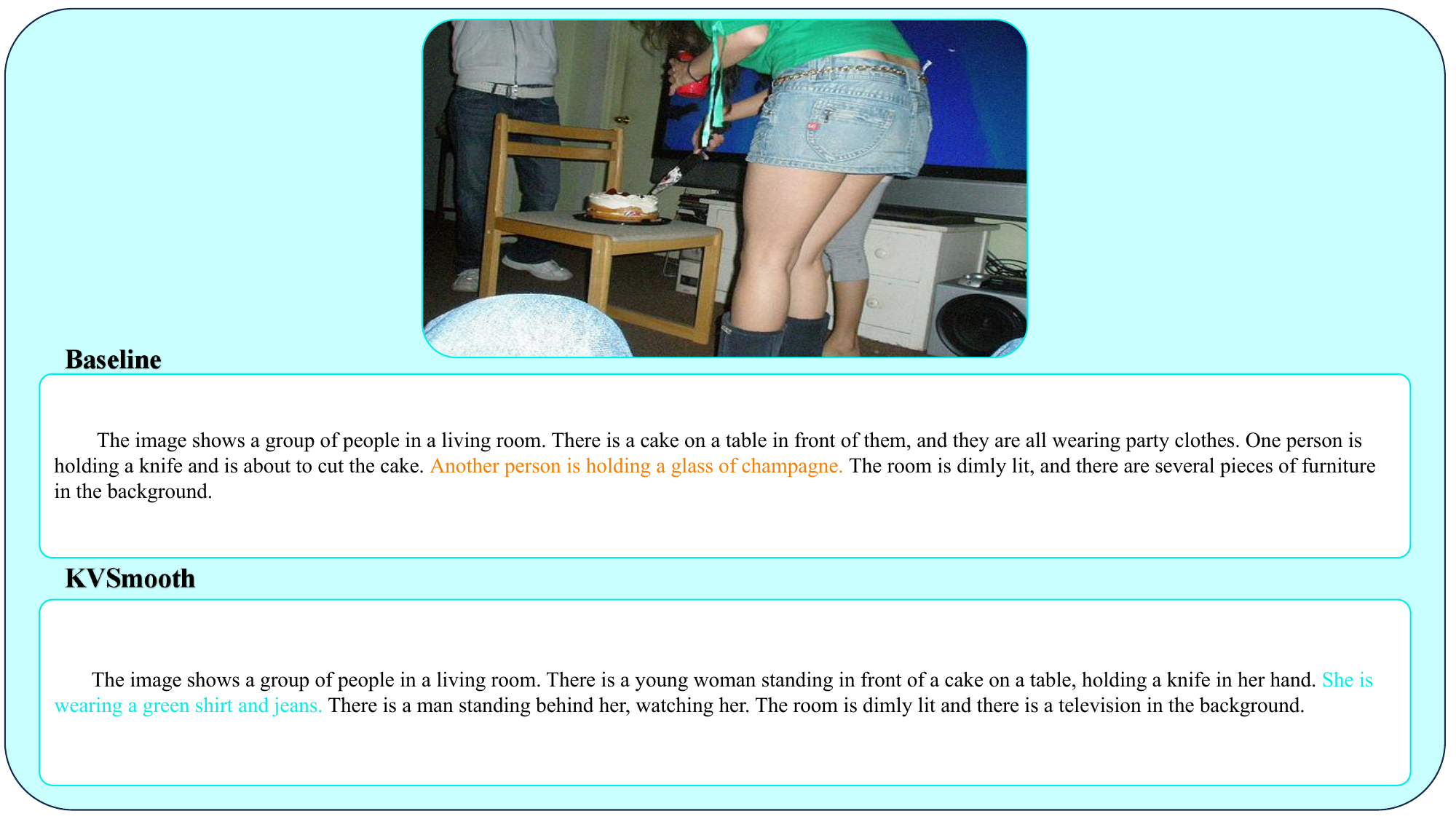}

\caption{Qualitative results on MiniGPT-4: baseline (w/o \name) vs.\ \name.}
\label{fig:qualitative_minigt4}
\end{figure*}
\begin{figure*}[t]
\centering

\includegraphics[height=0.31\textheight]{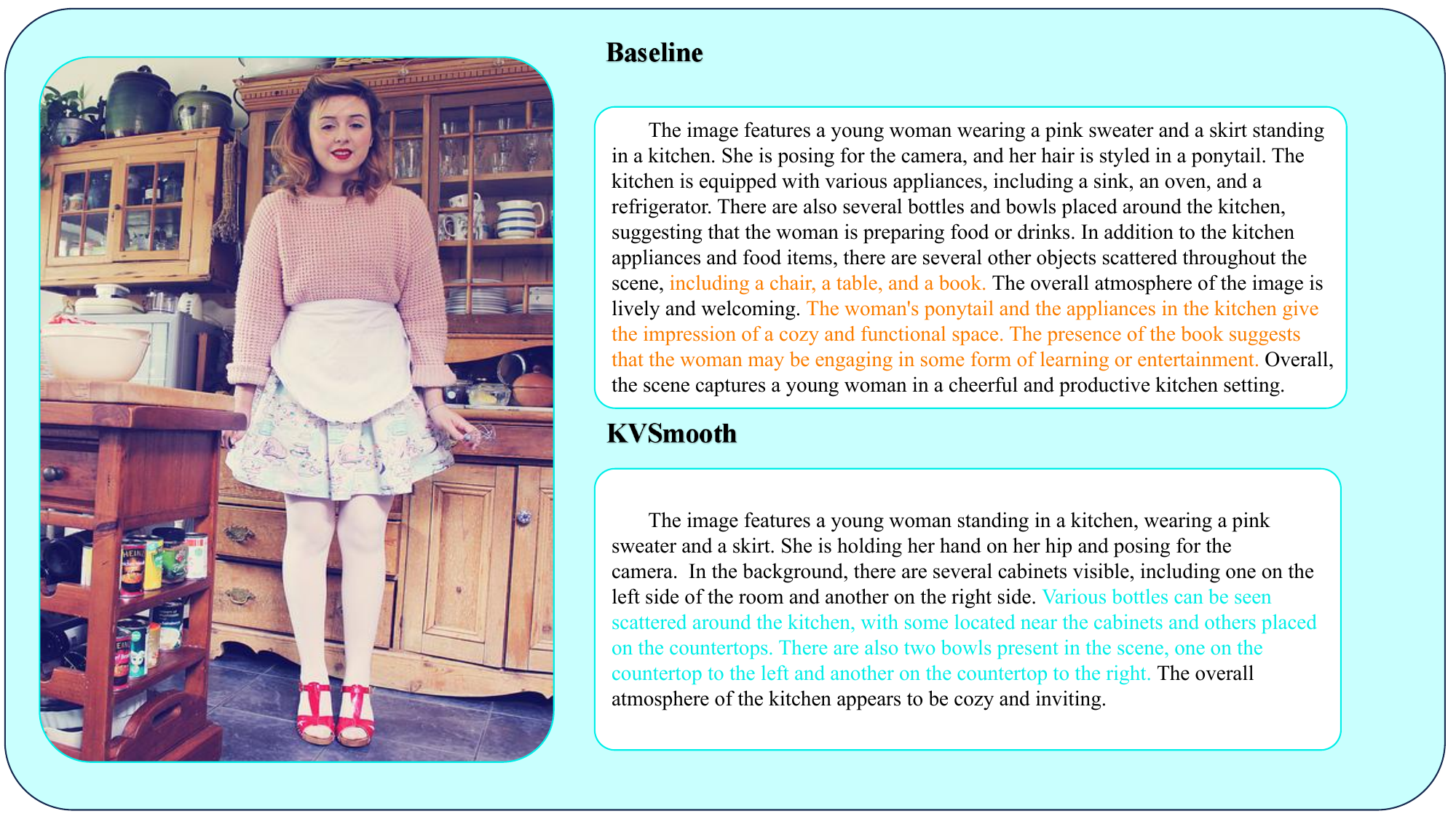}
\includegraphics[height=0.31\textheight]{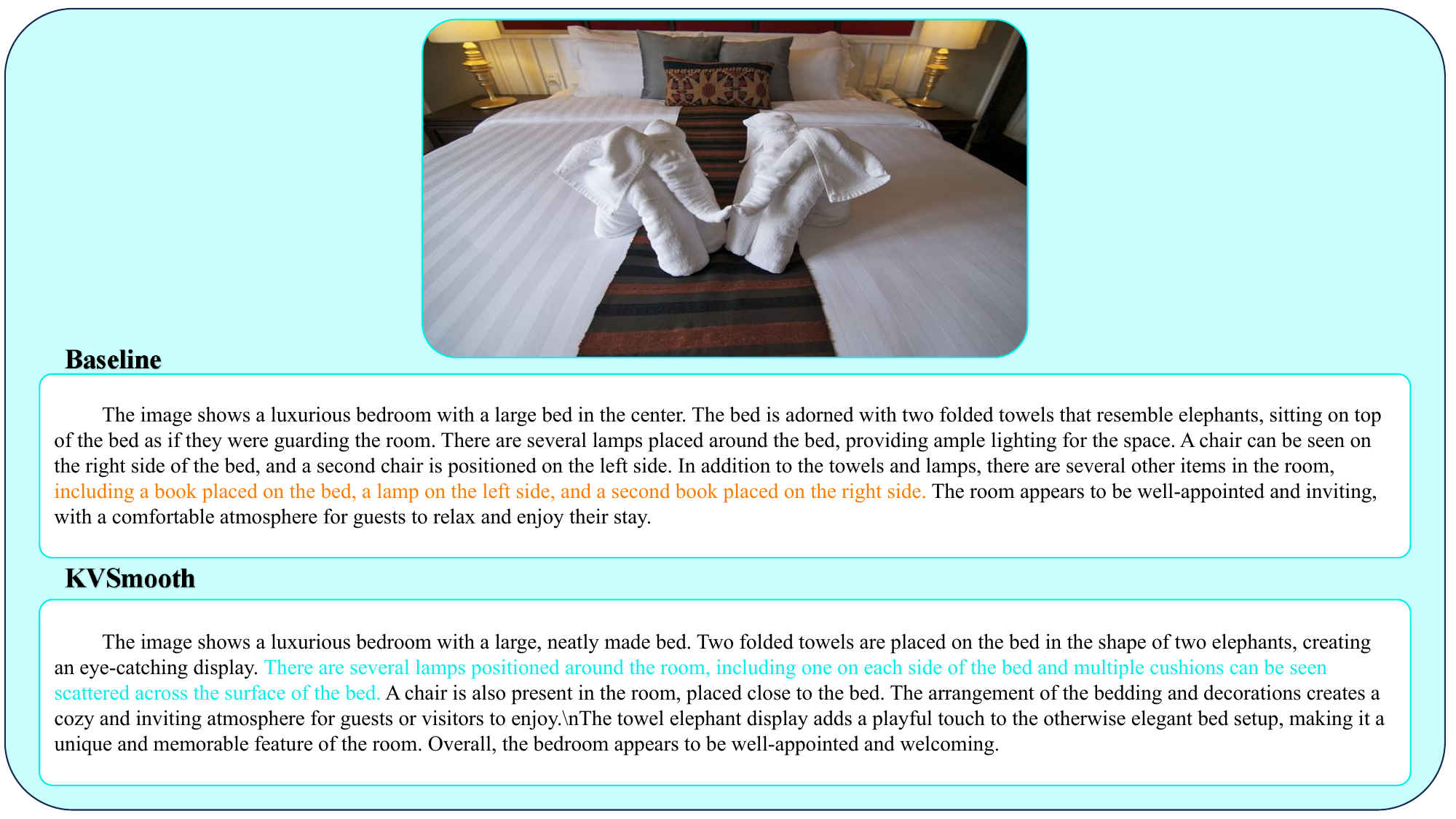}
\includegraphics[height=0.31\textheight]{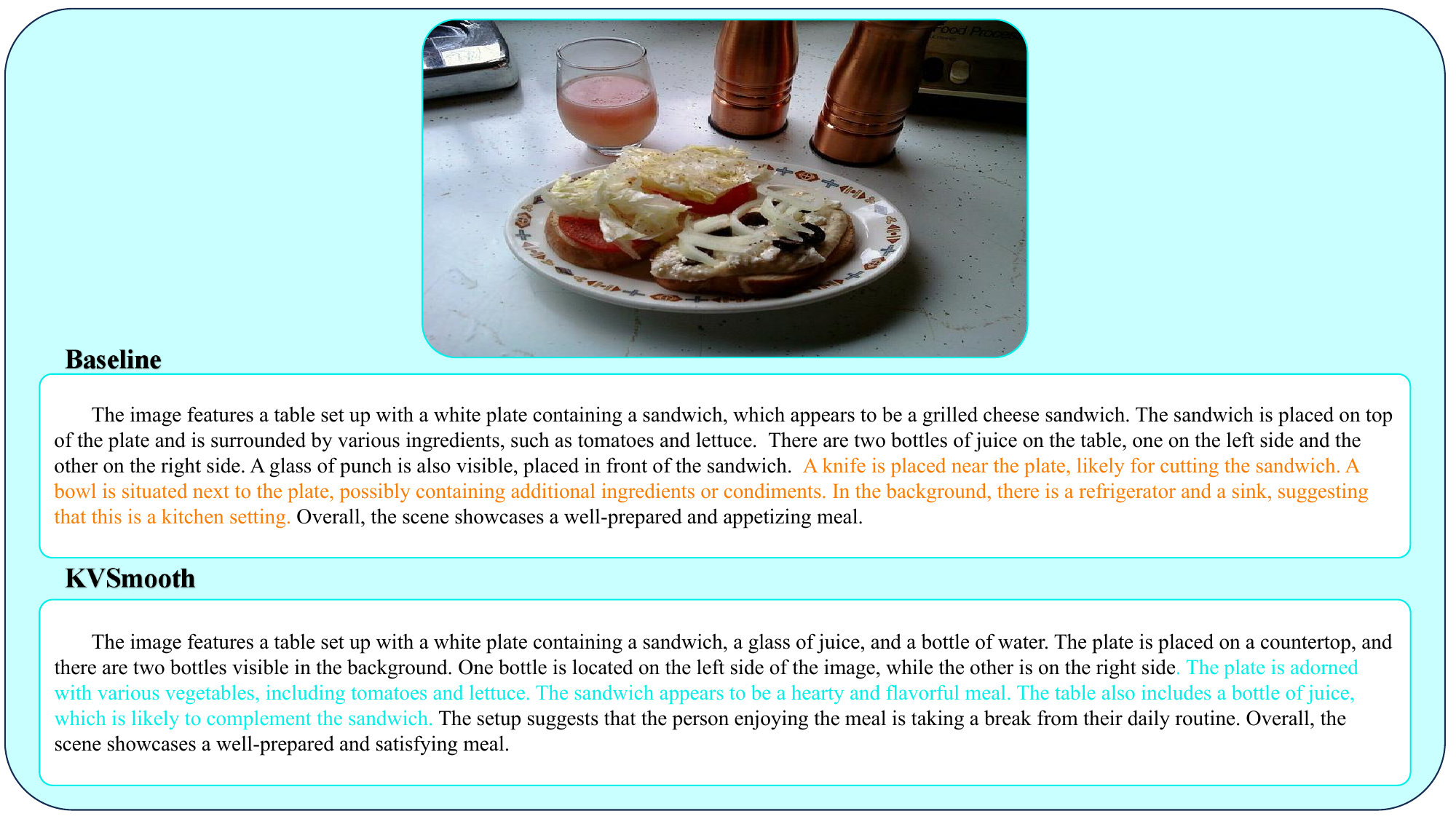}
\caption{Qualitative results on InstructBLIP: baseline (w/o \name) vs.\ \name.}
\label{fig:qualitative_blip}
\end{figure*}
\end{document}